\pdfoutput=1
\documentclass[10pt, logo, copyright]{nvidiatechreport}

\usepackage{mdframed}
\usepackage[utf8]{inputenc} %
\usepackage[T1]{fontenc}    %

\usepackage{amsfonts}       %
\usepackage{nicefrac}       %
\usepackage{microtype}      %
\usepackage[dvipsnames]{xcolor}         %
\usepackage{multirow}
\usepackage{multicol}
\usepackage{graphicx}
\usepackage[numbers]{natbib}
\usepackage{tabto}
\usepackage{xspace}
\usepackage{amsmath}
\usepackage{adjustbox}
\usepackage{enumitem}
\usepackage{wrapfig}
\usepackage{dblfloatfix}
\newcommand{\ours}{NLD-Uni\xspace}
\usepackage{float}

\usepackage{natbib}

\usepackage{hyperref}
\usepackage{url}
\usepackage{booktabs}
\usepackage{multirow}
\usepackage{enumitem}
\usepackage{adjustbox}
\usepackage{tabularx}
\usepackage{arydshln}
\usepackage{wrapfig}

\usepackage{multirow}
\usepackage{colortbl}
\usepackage{amsmath}
\usepackage{makecell}
\usepackage{hhline}
\usepackage{array}
\usepackage{diagbox}
\usepackage{graphicx}
\usepackage{adjustbox}
\usepackage{subcaption}

\usepackage{algorithm}
\usepackage{svg}
\usepackage{algpseudocode}
\usepackage{array}
\usepackage{amssymb}
\usepackage{booktabs}
\usepackage{subcaption}
\usepackage{tcolorbox}
\newcolumntype{H}{>{\setbox0=\hbox\bgroup}c<{\egroup}@{}}
\usepackage{multirow}

\usepackage{listings}
\usepackage{minted}
\newcolumntype{H}{>{\setbox0=\hbox\bgroup}c<{\egroup}@{}}

\NewDocumentCommand{\mvk}
{ mO{} }{\textcolor{green}{\textsuperscript{\textit{matthijs}}\textsf{{\small[#1]}}}}

\NewDocumentCommand{\shizhe}
{ mO{} }{\textcolor{magenta}{\textsuperscript{\textit{shizhe}}\textsf{{\small[#1]}}}}

\NewDocumentCommand{\wb}
{ mO{} }{\textcolor{teal}{\textsuperscript{\textit{wonmin}}\textsf{{\small[#1]}}}}

\usepackage{fp} %

\usepackage{xcolor}
\definecolor{mygreen}{rgb}{0.7, 1.0, 0.7}

\definecolor{myblue}{rgb}{0.7, 0.8, 1.0}

\usepackage{pifont}
\newcommand\mynuma[1]{\ifcase#1 \or \ding{172}\or \ding{173}\or
  \ding{174}\or \ding{175}\or \ding{176}\or \ding{177}%
  \or \ding{178}\or \ding{179}\or \ding{180}\or \ding{181}\else *\fi\relax}

\newcommand\mynumb[1]{\ifcase#1 \or \ding{182}\or \ding{183}\or
  \ding{184}\or \ding{185}\or \ding{186}\or \ding{187}%
  \or \ding{188}\or \ding{189}\or \ding{190}\or \ding{191}\else *\fi\relax}

\title{\ours: Advancing Masked Discrete Diffusion for High-Resolution Image Synthesis}

\author{Shufan Li*, Hanrong Ye*, Greg Heinrich*, Yonggan Fu, Haotian Zhang, Sifei Liu, Aditya Grover, Pavlo Molchanov \\
$^*$Core Contribution \\
}
\correspondingauthor{X}

\begin{abstract}
We propose Nemotron-Labs-Diffusion-Uni (\ours), a unified discrete diffusion large language model~(DLLM) for visual understanding and generation tasks. Compared with prior unified diffusion models, \ours~introduces two major innovations. First, we propose a unified cross-paradigm on-policy distillation (OPD) framework that successfully transfers the visual understanding ability of autoregressive VLMs and visual generation ability of continuous flow models, to a unified DLLM. 
Second, \ours incorporates a novel progressive self-refining diffusion process that gradually corrupts each token across multiple steps during training, as opposed to adopting the conventional masked diffusion formulation that directly corrupts a clean image token to an uninformative mask token. At inference, this design allows \ours~to make intermediate commit to a token, enabling progressive self-refinement for better visual quality. 
Through extensive experiments, we show that \ours~achieves state-of-the-art performance among unified DLLMs on diverse tasks including visual question-answering, multi-modal reasoning, object grounding, text-to-image generation, and image editing.  \href{https://huggingface.co/nvidia/NLD-Uni}{Model} and \href{https://github.com/NVlabs/Nemotron-Labs-Diffusion-Uni}{Code} are publicly available.

\end{abstract}

\newcommand{\greg}[1]{}
\newcommand{\PM}[1]{}

\newcommand{\ye}[1]{}

\title{Nemotron-Labs-Diffusion-Uni: Elastic Self-Refining Unified Diffusion LLM with Cross-Paradigm On-Policy Distillation}

\begin{document}
\maketitle

\begin{figure}[h]
    \centering
    \includegraphics[width=1.0\linewidth]{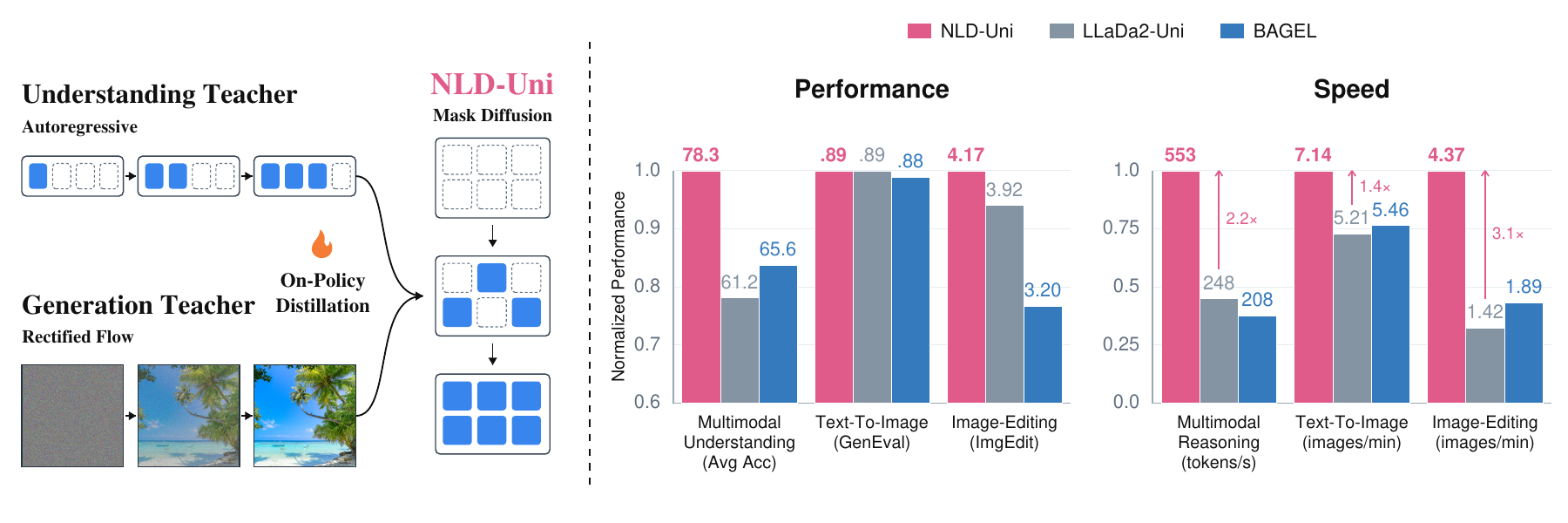}
    \caption{
    {(Left)} \ours introduces a new paradigm for building unified Diffusion LLMs through on-policy distillation from cross-paradigm teachers: VLMs for visual understanding and continuous diffusion models for visual generation. 
    {(Right)} \ours achieves highly competitive visual understanding and generation performance. Visual understanding performance is averaged across tasks in Table~\ref{tab:integrated-understanding-results}. Text-to-image and image editing reports GenEval and ImgEdit Score. Multimodal reasoning is measured on MathVista. Axes are normalized by the best model's performance on each task.
    }
    \label{fig:teaser}
\end{figure}

\section{Introduction}
\label{sec:intro}

Unified Multimodal Large Language Models (MLLMs) \citep{openai2024gpt4o,team2024gemini,chen2025blip3,chen2025janus} have achieved strong performance on diverse multimodal understanding and generation tasks. Compared with specialist models, they learn to perform diverse tasks via a unified sequence modeling interface, allowing synergetic learning that benefits each individual tasks~\citep{deng2025emerging}. They generally fall into two categories. The first design represents text and image as discrete tokens and employs a unified autoregressive (AR) modeling paradigm \citep{lu2022unified,chen2025janus,wang2024emu3}. The second design combines autoregressive modeling for text generation and continuous diffusion for image generation~(AR+Diff) \citep{deng2025emerging,pan2025transfer,mo2025x,nvidia2026cosmos3}. Several recent works such as LaViDa-O \citep{li2025lavidao} and MMaDA~\citep{yang2025mmada} explored a third alternative that builds a unified discrete diffusion language model~(unified DLLM) with masked diffusion paradigm (MDM). Unlike AR models or continuous flow models, MDMs randomly mask discrete tokens in a sequence and learn to predict the masked tokens during training. At inference, MDMs initialize a fully masked sequence and gradually unmask tokens across multiple diffusion steps. Compared with AR models, MDMs offer faster inference speed because they can generate multiple tokens in parallel \citep{sahoo2024simple,shi2024simplified}. Compared with AR+Diff setup, which needs to carefully balance conflicting AR and Diffusion objectives \citep{deng2025emerging}, unified DLLMs offer a single objective for both text and image, resulting in simplicity and, more importantly, scalability.

Despite these advantages, a substantial performance gap remains between existing unified DLLMs and domain-specialized VLMs and continuous flow models \citep{yang2025mmada,li2025lavidao}. In this work, we seek to narrow these gaps by introducing \ours, a unified DLLM featuring a new training paradigm and new architecture.

First, although on-policy distillation (OPD) has proven highly effective in improving the visual understanding capabilities of VLMs \citep{liu2026visual} and the visual generation capabilities of continuous flow models \citep{li2026diffusionopd}, adapting it to unified MDMs is challenging because of a lack of strong-performing MDM teachers. Distilling from strong AR VLMs and continuous flow models is not feasible because they follow fundamentally different paradigms. To overcome this challenge, we propose \textbf{cross-paradigm on-policy distillation (OPD)}, which enables \ours\ to learn from teachers with different generation paradigms, including an AR VLM teacher for visual understanding and a continuous flow teacher for visual generation.

\begin{figure}
    \centering
    \includegraphics[width=1.0\linewidth]{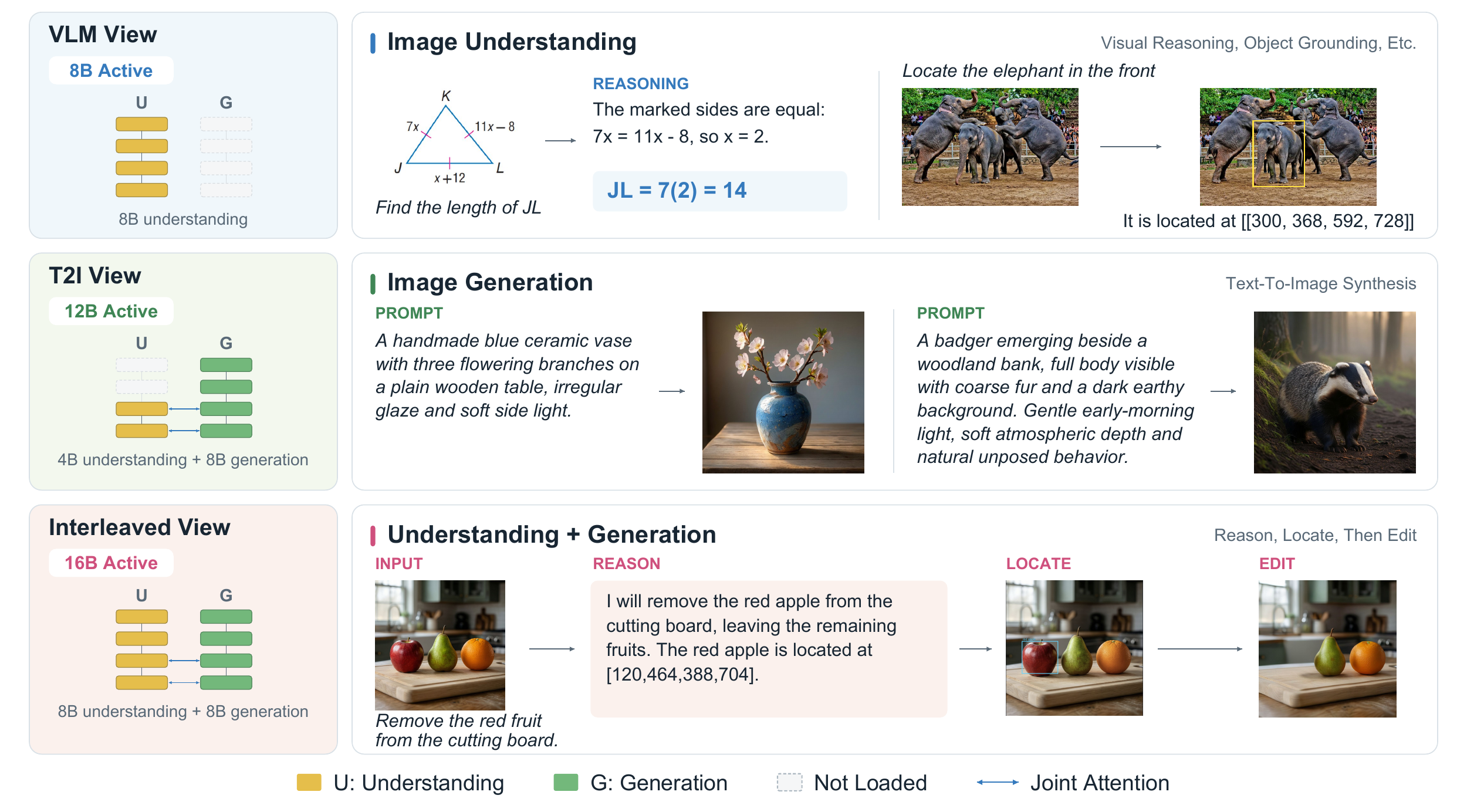}
    \caption{\textbf{\ours supports diverse multimodal tasks} by adopting an Elastic-MoT architecture with  three elastic views of a single model. The VLM view uses the 8B understanding branch; the T2I view combines 4B understanding parameters with the 8B generation branch; and the interleaved view uses both full branches (16B). Each view loads only the parameters needed for its task. Yellow and green blocks denote understanding and generation layers, gray blocks indicate omitted layers, and blue arrows denote joint attention. Representative examples demonstrate visual reasoning, object grounding, text-to-image synthesis, and image editing guided by reasoning and localization.}
    \label{fig:elastic_mot}
\end{figure}

Second, for visual generation tasks specifically, MDMs lack the self-refinement capabilities of continuous flow models. At inference, each token jumps directly from a masked state to a clean state and stays unchanged afterwards. Since MDMs sample tokens independently from per-token logit prediction, all tokens sampled at the same diffusion step are assumed to be conditionally independent. This assumption rarely holds in practice since tokens from the same image are highly correlated. To address this limitation and enable iterative refinement, we introduce a \textbf{self-refining masked diffusion process} that incorporates new intermediate states between the clean state and mask state. By breaking down the one-step jump to a multi-step transition, \ours~enables fine-grained inter-token interaction and progressive self-correction, resulting in better visual quality.

To evaluate the effectiveness of \ours, we conduct extensive evaluations on a diverse range of multimodal understanding and generation tasks, including visual question answering, object grounding, multimodal reasoning, text-to-image generation, and image editing. Experiments show that \ours~achieves state-of-the-art performance among unified DLLMs and greatly reduces the gap to specialized AR VLMs and continuous flow models.

The main contributions of this work are summarized as follows: 
1) We develop a cross-paradigm OPD framework that improves both visual understanding and generation by distilling knowledge from non-DLLM teachers.
2) We introduce a self-refining masked diffusion process that improves visual generation quality over the canonical masked diffusion process.
3) We present \ours, a state-of-the-art unified DLLM capable of both understanding and generating visual content.
We acknowledge that while extensive experiments show the effectiveness of our proposed methods, the model can be further improved via better data mixture and architecture optimization.

\section{Preliminaries and Related Works}

\subsection{Diffusion Language Models}
\label{sec:related_sbd}
Discrete diffusion language models \citep{lou2023discrete-sedd,sahoo2024simple,nie2025large} generate multiple text tokens in parallel, making them more efficient than AR models. However, its bidirectional attention introduces overhead compared to causal attention in AR models, particular for long sequences. Several works \citep{arriola2025block,cheng2026sdar} address this limitation through block-wise semi-auto-regressive generation. Notably, SBD \citep{gat2025set,zhu2026flare} incorporates variable-sized blocks during training, enabling flexible inference.

Formally, given a block of clean text tokens $B^0$, the forward diffusion process gradually mask it over the continuous time interval $t\sim[0,1]$ until it reaches a fully masked block $B^1$ containing only the mask token \texttt{[m]}. Given a multi-block token sequence $X=[B_1^0,B_2^0,...B_N^0]$, SBD models learn to predict the clean block $B_i^0$ from its partially masked version $B_i^t$ conditioned on all previous clean block. The optimization objective is defined as follows:
\begin{align}
    \mathcal{L}_\text{SBD}=\mathbb{E}_{i\sim \{1,2,..N\},t\sim[0,1],X\sim \mathcal{D}}[-\frac{1}{t}\log p(B_i^0|B_{j<i}^0,B_i^t)],
    \label{eq:sbd_loss}
\end{align}
where $\mathcal{D}$ is the data distribution. At inference, the blocks are sampled sequentially, but tokens in the same block are sampled in parallel across multiple diffusion steps. \ours~applies the SBD paradigm to unified multimodal modeling and treats each image as an integral block, whose size is resolution-dependent.  Note that when block size reduces to 1, SBD models reduce to an AR model.

\subsection{Masked Discrete Image Models}
Masked image models generate images by initializing a fully-masked image and gradually unmasking image patches. Each patch is encoded to a discrete code through vector quantization (VQ) \citep{esser2021taming}. MaskGiT \citep{chang2022maskgit} pioneered this paradigm. Messonic \citep{bai2024meissonic} scaled it to large-scale T2I. Recent works such as LaViDa-O and MMaDa \citep{li2025lavidao,yang2025mmada} explored integrating masked image modeling to unified MLLMs.  As discussed in Sec. \ref{sec:intro}, these models lack the self-refinement capabilities of continuous flow models. Some work mitigate this problem using a specialized unmasking schedule that prioritizes unmasking less-correlated tokens \citep{besnier2025halton,li2025lavidao}.  Others explored revising already unmasked \citep{wang2026remasking,li2026nemotron}. However, these works are sensitive to heuristic-based confidence thresholds and still present a mismatch between training and inference setup. 

\subsection{On-Policy Distillation (OPD)}

OPD \citep{agarwal2024gkd,gu2024minillm} enables capability transfers from a teacher model to a student model by providing teacher supervision signals on student-generated sequences. 
M-OPD~\citep{ma2026mopd} extends OPD to a multi-teacher setting to consolidate the capabilities of domain-specific experts. Several works~\citep{yuan2026vision,hao2026stabilizing} explored OPD for multimodal understanding tasks. Diffusion-OPD \citep{li2026diffusionopd} further applies it to continuous flow models for visual generation tasks. However, OPD for DLLMs is currently underexplored. OPDLM~\citep{su2026opdlm} is the most relevant work, but its scope is limited to same-model distillation that converts a pure AR model to a DLLM with weaker performance but faster inference. Unlike prior work, we propose cross-paradigm OPD to jointly distill knowledge from a strong AR VLM teacher and an image diffusion generation teacher into a unified DLLM.

Beyond the three areas above, \ours~is closely related to broader literature on unified MLLMs, and we provide more discussion in Appendix \ref{sec:related_unified}.

\section{Method}

\begin{figure}
    \centering
  \vspace{-30pt}
    \includegraphics[width=1.0\linewidth]{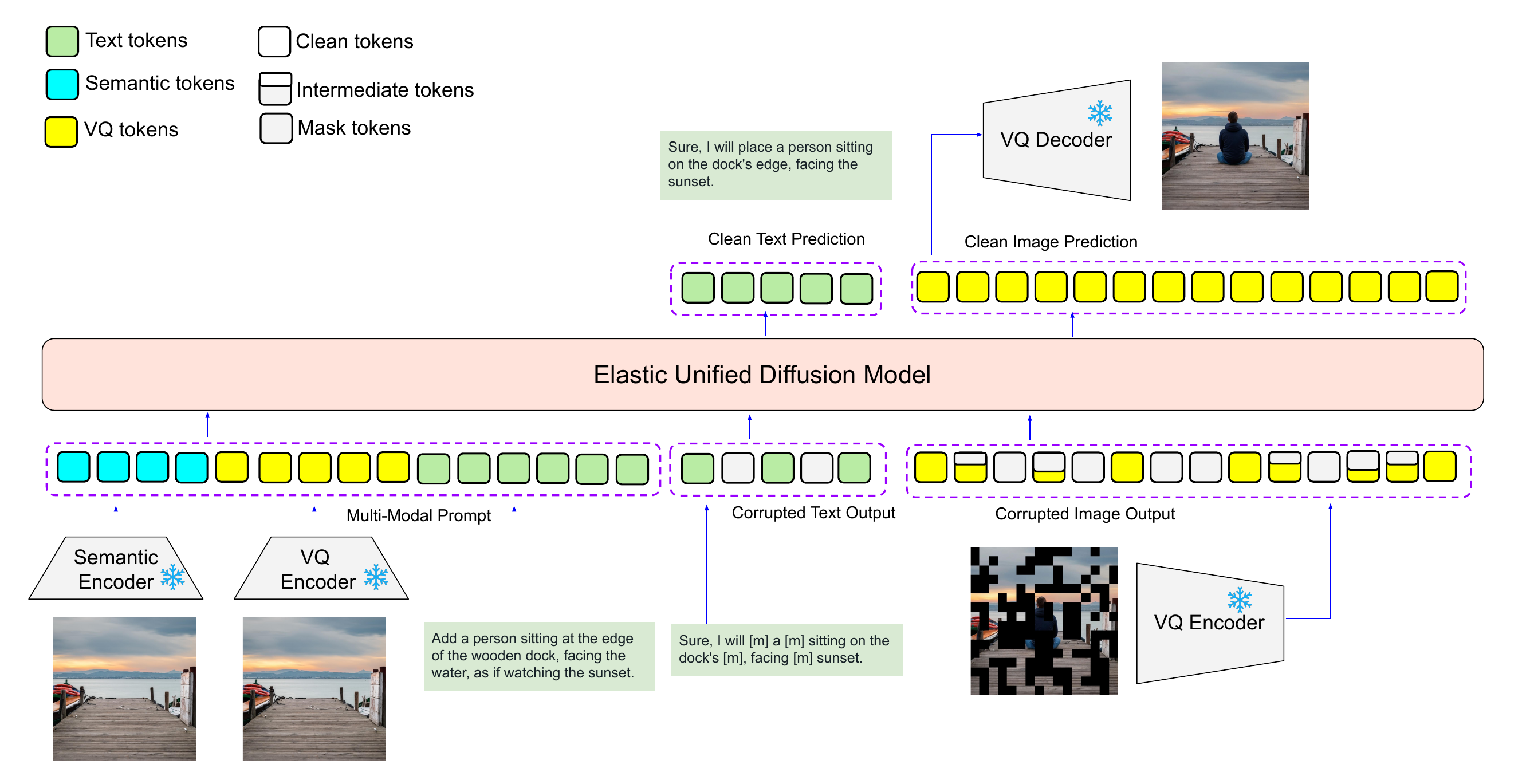}
    \caption{\textbf{Overall Framework of \ours.} \ours~is a unified diffusion model whose input and output are both multimodal. Images in the input stream are represented by both continuous embeddings of a semantic encoder and discrete tokens of a VQ encoder. Images in the output stream are represented solely using VQ codes. \ours~is trained to predict the clean token distribution given partially masked inputs. For image generation, we incorporate an intermediate masking state.}
    \label{fig:explainer}
\end{figure}
\subsection{Model Architecture}

\ours~adopts a two-tower Elastic Mixture-of-Transformer (Elastic MoT) design following Lavida-O \citep{li2025lavidao}, which consists of an understanding expert tower and generation expert tower. They interact with each other through a joint attention mechanism. The understanding tower is initialized with a diffusion language model NLD \citep{fu2026nemotron}, which uses the block-causal SBD formulation described in Section \ref{sec:related_sbd}. The generation tower is initialized from an identical copy. Following the common design of unified MLLMs \citep{deng2025emerging,li2025lavidao}, we encode images with both a semantic encoder and VQ encoder. Once images and texts are tokenized, the canonical SBD formulation can be directly applied. Elastic-MoT architecture is illustrated in Figure \ref{fig:elastic_mot}. Figure \ref{fig:explainer} illustrates the overall framework.

\subsection{Discrete Diffusion with Progressive Self-Refinement}
\label{sec:self_refinement}

The canonical mask discrete diffusion uses binary mask states, where a token directly jumps between the mask and clean states. Differently, our \ours~incorporates intermediate states using the concept of token age $a\in \mathcal{A}= \{0,1,2...K\}$. In particular, $a=0$ implies a token is in the clean state, $a=K$ implies a token is in the fully-masked state. Crucially, $a$ is a token-local discrete index and is different from the global continuous timestep $t\in[0,1]$. 

\begin{figure}
    \centering
    \includegraphics[width=1.0\linewidth]{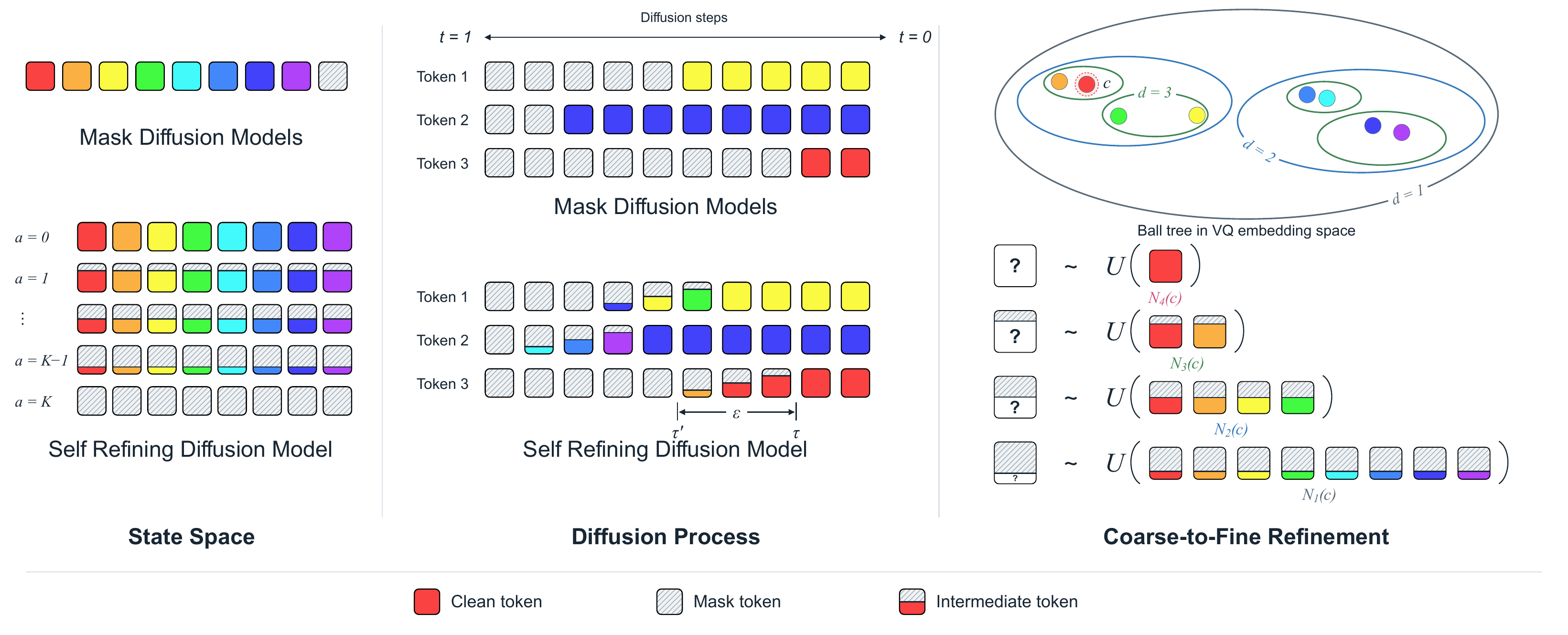}
    \caption{\textbf{Design of Self-Refinement Process.} (Left) We extend the state space of mask diffusion with token age $a$ with $a=0$ being the clean state and $a=K$ being the masked state. (Middle) Instead of one-step jump, each token is progressively masked through the interval $(\tau,\tau')$ with length $\epsilon$.  Colors encode different tokens. (Right) We implement the progression from age $a=0$ to $a=K$ through random jumps based on a ball-tree in the embedding space. Token distributions are diffused from a one-hot vector to balls with growing sizes. This example has tree depth $D=4$.}
    \label{fig:self_refinement_main_paper}
\end{figure}

Formally, given a vocabulary $\mathcal{V}$, we adopt the state space $S=\mathcal{V}  \times \mathcal{A}$ as opposed to $S=\mathcal{V}  \cup \{ [m]\}$ of standard MDMs. This is shown in Figure \ref{fig:self_refinement_main_paper} (Left). We define the self-refining diffusion process as the interaction of a \textbf{discrete time per-token corruption process} and a \textbf{continuous time per-token jump schedule}.  Specifically, each token in $(c,a)\in S$ can only transition to $(c', a+1)$ in the forward process, with $(*,0)$ being the initial states and $(*,K)$ being the terminal states.  The transition probability is defined as $U_{k|k-1}(u|c)$, which defines the probability of a token $(c,k-1)$ transits to $(c,k)$. This generates a discrete time per-token corruption process where each token undergoes $K$ jumps from initial clean state and terminal mask state. Given a token $X^0_i=(C_i^0,A_i^0)$, the continuous time per-token jump schedule $b_i^t(a)=\mathbb{P}(A_i^t=a)$ determines when the jumps occur for each token in the interval $[0,1]$. It satisfies $b_i^0(0)=1$ and $b_i^1(K)=1$ at the initial and terminal states.

\textbf{Per token Corruption Process.} The per-step transition rate defined above generates the marginal $U_{j|k}(u|c)$ with $0 \le k \le j \le K$. It must satisfy $U_{K|k}(u|c)=Q(u)$, where $Q(u)$ is a stationary distribution over $\mathcal{V}$, for all $k<K$ because the terminal state is semantically fully masked and cannot correlate to the initial states. We implement this by randomly replacing a token c with a sampled token u in the VQ embedding space, where the jump distance depends on the token age. Concretely, we construct a balanced binary ball tree \citep{huang2023lightweight} which is commonly used in efficient nearest neighbor search. Each node in the ball tree represents a high-dimensional ball. in the embedding space, with parent nodes having a larger radius than child nodes. Let $N_d(c)$ be the set of all tokens that share a common ancestor to $c$ at depth $d$, and $D$ be the tree depth, it satisfies $N_1(c)=\mathcal{V}$ for the root node and $N_{D}(c)=\{c\}$ for the leaf node. An example of this structure is shown in Figure \ref{fig:self_refinement_main_paper} (top-right), where we visualize a 4-depth tree. 

We define $U_{k|{k-1}}(u|c)$ as the uniform distribution over $N_{D-k}(c)$, which gives the following  forward and reverse marginal:
\begin{align}
    U_{j|k}(u|c)= \mathbf{1}\{u\in N_{D-j}(c) \}\frac{1}{|N_{D-j}(c)|};\quad  U_{k|j}(c|u,\hat{c})= U_{k|0}(c|\hat{c}), \forall u \in N_{D-j}(\hat{c})
    \label{eq:fwd_reverse_corruption}
\end{align}
where $\hat{c}$ is the initial token state at $a=0$, and $0\le k < j \le K$. LHS satisfies that the initial distribution is a one-hot distribution of the clean token and the terminal distribution is the uniform distribution over all masked states $(*,K)$.  RHS can be used at inference by substituting the clean token $\hat{c}$ with a model prediction.  We defer full derivation to Appendix \ref{sec:appendix_ball_tree}. Figure \ref{fig:self_refinement_main_paper} (bottom-right), illustrates the progression of marginal $U_{j|0}(\hat{c})$. It corrupts the token $\hat{c}$ to progressively random tokens in progressively larger balls in the embedding space, until it reaches the terminal state.  We adopt the ball-tree mainly for efficiency. The full pairwise distance matrix would require 32GB VRAM since \ours has 131k distinct VQ codes. By contrast,  the corresponding 18-depth binary tree can be represented via a single 1D array taking only 524 KB VRAM.

\textbf{Per-Token Jump Schedule} embeds the discrete jumps in the continuous interval $[0,1]$. We adopted DNDM \citep{chen2024fast}'s setup which generates the same forward marginal as vanilla MDM by first sampling a per-token jump time $\tau_i\sim U([0,1])$ for each token $i$. The forward token-age transition then becomes deterministic with each token jumps from the clean state to the mask state exactly at $t=\tau_i$. We extend this to support multiple intermediate states by sampling the interval $(\tau_i,\tau_i')\subset[0,1]$ via $\tau_i\sim U ([0,1-\epsilon])$ and set $\tau_i'=\tau_i+\epsilon$, where $\epsilon$ is a hyperparameter controlling the duration of intermediate states. When $\epsilon\rightarrow0$, the process reduces to DNDM. When $\epsilon=1$, all tokens are progressively corrupted at the same rate. To minimize the complexity, we let all intermediate token ages $1,2,..K-1$ to progress uniformly in $(\tau,\tau')$, taking $\frac{\epsilon}{K}$ time each. Given fixed $\epsilon\in(0,1)$, this establishes a bijection between every instance of per-token jump schedule to a one-step jump schedule in DNDM. This process becomes non-Markov as we cannot marginalize $\tau_i$. At training and inference, we presample $(\tau_i,\tau_i')$ for each token, then the changes in token age become deterministic. The forward and reverse marginal can be obtained by directly applying Equation \ref{eq:fwd_reverse_corruption} with change-of-variable. Particularly, at inference, token $i$ stays masked in $[\tau'_i,1]$, is progressively refined in $[\tau_i,\tau_i']$ and stays fixed in $[0,\tau_i]$. We defer additional details and derivations to Appendix~\ref{sec:appendix_jump_schedule}. 

The most relevant work to \ours~in this aspect is MDM-prime \citep{chao2026beyond}, which breaks tokens into subtokens to enable partial masking. However, its image experiments are limited to 32x32 low-res images. We find that it does not scale to our setting. By contrast, \ours~makes use of codebook geometry to enable partial masking and unmasking in a coarse-to-fine manner, without the need to learn explicit sub-token representations. We provide more discussion in Appendix \ref{sec:appendix_mdm_prime}.

\begin{figure}
    \centering
    \includegraphics[width=1.0\linewidth]{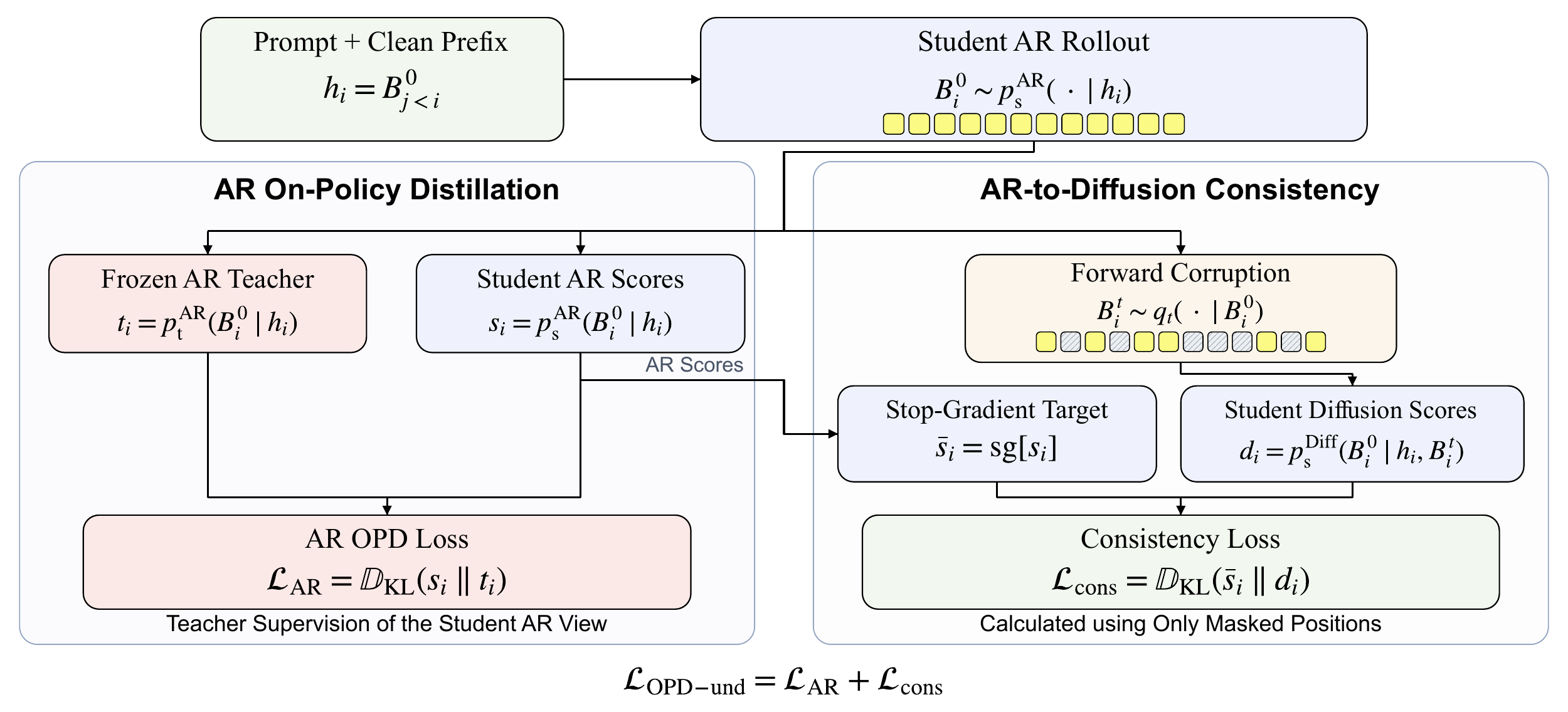}
    \caption{\textbf{On-Policy Distilling from Autoregressive VLMs.} Top: Given a prompt and clean prefix blocks, we generate the current block with the student's AR view by temporarily adopting a block size of 1. Left: AR on-policy distillation loss aligns the student’s AR distribution with a frozen AR teacher. Right:  Consistency loss aligns the student's diffusion view to its own AR view. The diffusion prediction is obtained by evaluating the student on randomly masked AR rollouts. We apply stop-grad to the  AR target.}
    \label{fig:distillation_und_math}
\end{figure}

\subsection{Distillation of Visual Understanding Ability}
\label{sec:distill_ar}

To distill knowledge from AR teachers, we exploit SBD's ability to operate with varying block sizes. As described in Section~\ref{sec:related_sbd}, SBD models the conditional distribution $p(B_i^0 \mid B_1^0,\ldots,B_{i-1}^0,B_i^t)$ for a sequence of blocks $X=[B_1^0,\ldots,B_N^0]$, predicting each clean block from its corrupted version and the preceding clean blocks.
When block size is one, SBD reduces to a standard AR model (AR-view). Hence, we can directly supervise the student's AR view with an AR teacher on student's AR rollouts. However, using AR OPD loss alone would lose the capability of generating multiple tokens without any supervision on the diffusion view. 

A naive design is to directly copy the teacher score to the diffusion following OPD-LM \citep{su2026opdlm}, however it creates ill-formed supervison signals due to context mismatch. As a concrete example, consider the student answer \texttt{"Jason visited San Francisco"} for a QA task. A teacher may want to respond \texttt{"Jason visited New York"} conditioned on the first two words and assigns high confidence to the token \texttt{"New"}. This supervision target is reasonable in AR setting. However, if we employ a block size of 2 and consider the partial state \texttt{"[m] Francisco"}, directly copying scores of AR teacher would guide the student generation towards generating \texttt{"New Francisco"}, which is ill-formed. 

To address this challenge, we propose a transitive distillation objective that supervises the student's AR view with teacher scores and supervises the student's diffusion view with its own AR view. Specifically, we first generate a clean block $B_i^0$ and supervise it with AR OPD loss. To supervise the diffusion view, we apply the forward process and corrupt $B_i^0$ to a partially masked version $B_i^t$ and minimize the KL divergence between the student's AR prediction and diffusion prediction. The final understanding loss can be defined as follows:
\begin{align}
    \mathcal{L}_{\text{OPD-und}} = \underbrace{\mathbb{D}_{\text{KL}}(p_{\text{s}}^{\text{AR}}(B_i^0|B_{j<i}^0) \parallel p_{\text{t}}^{\text{AR}}(B_i^0|B_{j<i}^0))}_{\text{AR OPD Loss}} + \underbrace{\mathbb{D}_{\text{KL}}(p_{\text{s}}^{\text{AR}}(B_i^0|B_{j<i}^0) \parallel p_{\text{s}}^{\text{Diff}}(B_i^0|B_{j<i}^0,B_i^t))}_{\text{Consistency Loss}}
\end{align}
where $p_s,p_t$ denote student and teacher distributions. The first term is the standard AR OPD loss, the second term is a self-consistency loss. We apply stop-grad for $p_s^{AR}$ in consistency loss and only evaluate masked positions. Crucially, while the context is still different, the AR scores have good alignment because tokens in $B_i^0$ are sampled from $p_s^{AR}$ with a low temperature.\footnote{We provide studies on sampling temperature in Appendix \ref{sec:appendix_ablation_opd_und}} We note that OPD-LM~\citep{su2026opdlm} is applied to AR2Diff conversion where the student is initialized with the teacher weights. The teacher's gap to the student's AR-view is initially 0.  Our setup distills an AR teacher that is larger and stronger to an already trained student diffusion model, which is more general but has a considerable teacher-student distribution gap at initialization, creating ill-formed signals when applying OPD naively.\footnote{OPD-LM also finds that for Qwen-8B,  self-supervision outperforms Qwen-32B teacher in AR2Diff.} The overall pipeline is illustrated in Figure \ref{fig:distillation_und_math}.
 We defer additional details to Appendix \ref{sec:appendix_ablation_opd_und}.
\begin{figure}[t]
    \centering
    \includegraphics[width=1.0\linewidth]{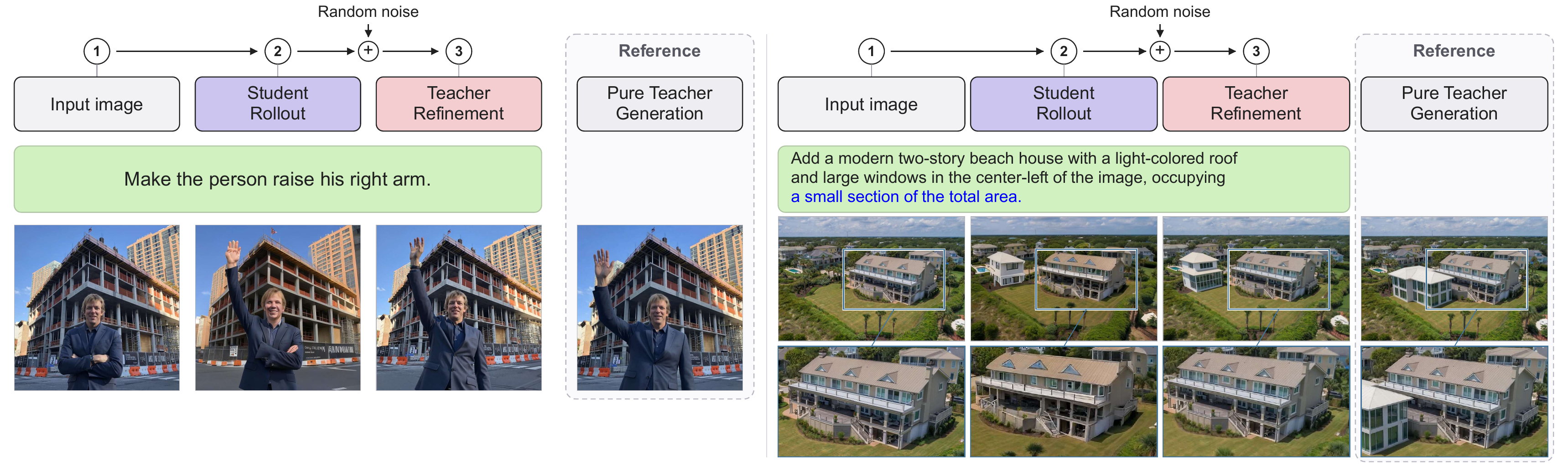}
    \caption{\textbf{On-Policy Distilling from Continuous Flow Models.} 
    Given a student output, we generate a teacher-refined output by adding noise to the student sample and re-denoise it with the teacher model. For reference, we also show an independent teacher generation. We observe that the teacher's refinements are closely aligned to the student's output while having better visual fidelity and source image preservation.  }
    \label{fig:distillaion}
\end{figure}

\subsection{Distillation of Visual Generation Ability}

Distilling continuous flow models to DLLM is more challenging as flow models operate in continuous latent spaces, making exact KL-style distribution matching infeasible. However, the general notion of using teacher supervision on student-generated sequences is still feasible. In this work, we explore cross-paradigm OPD for image editing tasks with Qwen-Image \citep{wu2025qwen} teacher. Given a student output, we first re-encode it with the teacher's continuous VAE. We then create a teacher supervision target by first adding noise to the student generation in teacher's latent space and re-denoise it with the teacher model, conditioned on the input image and instruction. We call this target teacher-refined student output. This pipeline is shown in Figure \ref{fig:distillaion} (right). We observe that teacher-refined images are semantically and spatially aligned to student generation, but with reduced artifacts, higher fidelity, and better source-image preservation. By contrast, independent teacher generations exhibit a larger gap to student generation. This distinction makes the refined samples ideal candidates for training targets. We also found that teacher-refined outputs can be better than independent teacher generations where the teacher model makes mistakes while the student model does not. We defer more discussion and examples to Appendix \ref{sec:opd_cont_appendix_method}.  

We adopt the canonical SFT loss on the teacher-refined samples. While this is not ``on-policy'' in the conventional sense, it constructs teacher-refined targets from the current student's own rollouts, adapting supervision to its evolving generation distribution. It closely connects to several concurrent OPD studies in AR  language modeling \citep{jiang2026trajectory}, which also explored using teacher models to correct student rollouts.

\section{Experiment results}

\subsection{Main Results}

\textbf{Setup.} We train \ours~in a three-stage manner. The first stage primarily includes text-image pairs for basic visual alignment and image generation capabilities. The second stage introduces additional data for instruction following, image editing, object grounding, and interleaved generation. The final stage is the unified OPD training stage, with Nemotron-3 Nano Omni as the teacher for multimodal reasoning tasks and Qwen-Image for image editing tasks. We also mix T2I SFT batches to preserve the text-to-image generation performance. Notably, we do not include an OPD teacher for T2I tasks because the base model's performance on GenEval and HPSv3 is stronger than teacher candidates like Qwen-Image and Flux. The training is conducted on 64 H100 GPUs over 26 days. In particular, the final OPD stage takes 5 days. We defer additional details to Appendix \ref{sec:appendix_training_setup}.

\textbf{Generic Understanding Tasks.} We report the performance of image understanding tasks in Table \ref{tab:integrated-understanding-results}. \ours outperforms the previous state-of-the-art unified diffusion model and diffusion VLMs like LLaDa-V \citep{you2025lladav}, NLD-VLM \citep{fu2026nemotron} and LaViDa-O \citep{li2025lavidao} by considerable margin on a wide range of benchmarks such as MMMU \citep{yue2023mmmu}, ChartQA \citep{masry-etal-2022-chartqa}, InfoVQA~\citep{mathew2022infographicvqa} and AI2D\citep{kembhavi2016diagram}. On object grounding benchmark RefCOCO \citep{kazemzadeh2014referitgame} and GUI grounding benchmark ScreenSpot-Pro \citep{li2025screenspot}, \ours~also demonstrates strong performance while most unified models are not equipped with this ability, unlocking new potential applications for diffusion language models. These gains are mostly attributed to scaling training data.  Overall,  \ours  reduces the gap between unified diffusion models and state-to-the-art AR models on visual understanding.

\textbf{Multimodal Reasoning Tasks.}  We report the performance of multimodal reasoning tasks in Table \ref{tab:math-reasoning-results} and compare \ours~with pre-OPD checkpoint (\ours-Base). Our proposed cross-paradigm OPD framework improves performance on a wide range of tasks, with most notable improvements in We-Math \citep{qiao2025we}, Geometry3k \citep{lu2021inter}, and MathVerse \citep{zhang2024mathverse}. It also achieves state-of-the-art performance among diffusion language models. We additionally include a naive OPD baseline using OPD-LM \citep{su2026opdlm} style loss which directly copies teacher scores as the supervision target for diffusion generation, this worsens the performance of \ours-Base because of the mismatch we identified in Section \ref{sec:distill_ar}. These results highlight the effectiveness and necessity of the proposed cross-paradigm OPD framework.

\begin{table*}[t]

\centering
\caption{\textbf{Quantitative results on image-level understanding tasks.} \ours~is evaluated using a confidence-based diffusion sampler following \cite{wu2025fast}. }
\label{tab:integrated-understanding-results}
\resizebox{\textwidth}{!}{
\scriptsize
\setlength{\tabcolsep}{3pt}
\begin{tabular}{l*{12}{c}H}
\hline
\textbf{Model}
& \textbf{MMMU}
& \textbf{RealWorldQA}
& \textbf{ChartQA}
& \textbf{DocVQA}
& \textbf{InfoVQA}
& \textbf{TextVQA}
& \textbf{AI2D}
& \textbf{RefCOCO}
& \textbf{Sci.QA}
& \textbf{ScreenSpot-Pro}
& \textbf{OCRBenchV2}
& \multirow{2}{*}{\textbf{Avg}}
& \textbf{Speed} \\
\textit{Split}
& val & test & test & val & val & val
& test & val & test (img) & test & test (EN) & & task/sec \\
\hline
\multicolumn{14}{c}{\textit{Autoregressive Models}} \\
LLaVA-1.6-7B \cite{liu2024llavanext}
& 35.1 & - & 64.9 & 74.4 & 37.1 & - & 66.6 & - & 73.2 & - & - & - & - \\
Qwen2.5-VL-7B \cite{bai2025qwen25-vl}
& 58.6 & - & 84.9 & 82.6 & - & - & 83.9 & - & - & - & - & - & - \\
Qwen3-VL-8B~\cite{bai2025qwen3}
& 69.6 & 71.5 & 89.6 & 95.6 & 83.4 & 82.9 & 84.8 & 87.3 & 75.2 & 54.6 & 64.1 & 78.1 & 16.5 \\
Cosmos3-Nano~\cite{nvidia2026cosmos3}
& 60.9 & 72.2 & 85.5 & 94.5 & 81.5 & 83.4 & 85.0 & 79.2 & 93.0 & 56.1 & 65.9 & 77.9 & 31.3 \\
BAGEL \cite{deng2025emerging}
& 55.3 & 73.9 & 74.3 & 94.3 & 60.7 & 86.0 & 88.9 & 33.3 & 97.3 & 3.4 & 53.9 & 65.6 & 25.4 \\
\hline
\multicolumn{14}{c}{\textit{Discrete Diffusion Models}} \\
Dimple \cite{yu2025dimple}
& 45.2 & - & 63.4 & - & - & - & 74.4 & - & 77.1 & - & - & - & - \\
LaViDa \cite{li2025lavida}
& 43.6 & 54.5 & 64.6 & 59.0 & 34.2 & - & 70.0 & - & 80.2 & - & - & - & - \\
NLD-VLM~\cite{fu2026nemotron}
& 50.4 & 60.3 & 76.6 & 88.3 & 57.8 & 71.8 & 74.7 & 51.6 & 72.0 & 0.1 & 37.5 & 58.3 & 36.5 \\
Fast-DVLM~\cite{wu2026fast}
& 44.6 & 65.1 & 82.8 & 92.1 & 75.6 & 76.1 & 79.7 & 35.6 & 77.0 & 0.0 & 38.9 & 60.7 & 15.1 \\
LLaDa-V \cite{you2025lladav}
& 48.6 & 63.2 & 78.3 & 83.9 & 66.3 & - & 77.8 & - & 0.0 & - & - & - & - \\
MMaDa \cite{yang2025mmada}
& 30.2 & - & 9.8 & 10.9 & 14.9 & - & 66.6 & - & 55.8 & - & - & - & - \\
LaViDa-O~\cite{li2025lavidao}
& 45.1 & - & 80.0 & 73.7 & 44.6 & - & 76.7 & 91.2 & 84.6 & 0.0 & - & - & - \\
LLaDA2-Uni~\cite{ai2026llada2}
& 50.1 & 66.7 & 80.1 & 89.5 & 70.1 & 70.0 & 82.0 & 39.9 & 88.2 & 0.4 & 36.0 & 61.2 & 9.2 \\
\rowcolor{cyan!10}
\ours
& 60.7 & 70.5 & 89.9 & 94.2 & 84.0 & 76.2
& 85.0 & 94.1 & 95.4 & 49.2 & 62.1 & {78.3} & 71.4 \\
\rowcolor{cyan!10}
$\Delta$ vs.\ Diff. SOTA
& \textcolor{green!60!black}{\textbf{\textit{+10.3}}}
& \textcolor{green!60!black}{\textbf{\textit{+3.8}}}
& \textcolor{green!60!black}{\textbf{\textit{+7.1}}}
& \textcolor{green!60!black}{\textbf{\textit{+2.1}}}
& \textcolor{green!60!black}{\textbf{\textit{+8.4}}}
& \textcolor{green!60!black}{\textbf{\textit{+0.1}}}
& \textcolor{green!60!black}{\textbf{\textit{+3.0}}}
& \textcolor{green!60!black}{\textbf{\textit{+2.9}}}
& \textcolor{green!60!black}{\textbf{\textit{+7.2}}}
& \textcolor{green!60!black}{\textbf{\textit{+48.8}}}
& \textcolor{green!60!black}{\textbf{\textit{+23.2}}}
& \textcolor{green!60!black}{\textbf{\textit{+17.1}}}
& \textcolor{green!60!black}{\textbf{\textit{+34.9}}} \\
\hline
\end{tabular}
}
\end{table*}

\begin{table*}[t]
\centering
  \vspace{-5pt}
\caption{\textbf{Quantitative results on multi-modal reasoning benchmarks.} *Denotes naively applying teacher score to student's diffusion view using OPD-LM's loss.
}
\label{tab:math-reasoning-results}

\resizebox{\textwidth}{!}{
\scriptsize
\setlength{\tabcolsep}{4pt}
\begin{tabular}{lccccccc}
\hline
\textbf{Model}
& \textbf{MathVista}
& \textbf{DynaMath}
& \textbf{MathVision}
& \textbf{MathVerse}
& \textbf{Geometry3k}
& \textbf{LogicVista}
& \textbf{We-Math} \\
\hline
\multicolumn{8}{c}{\textit{Autoregressive Models}} \\
Qwen2.5-VL-7B \cite{bai2025qwen25-vl}
& 68.2 & 55.7  & 22.4 & 39.0 & 37.1 & 42.4 & 33.3 \\
Qwen3-VL-8B \cite{bai2025qwen3}
& 77.2 & 62.2 & 37.9 & 67.7 & 57.5 & 50.1 & 69.0 \\
BAGEL \cite{deng2025emerging}
& 73.1 & - & 24.1 & - & - & - & 45.8 \\
\midrule
\multicolumn{8}{c}{\textit{Discrete Diffusion Models}} \\

Fast-dVLM \cite{wu2026fast}& 57.3 & 51.3 & 20.2 & 33.7 & 27.0 & 37.1 & 41.0 \\
NLD-VLM \cite{fu2026nemotron}  & 56.0 & 48.2 & 21.5 & 28.5 & 29.3 & 38.3 & 50.1 \\
LLaDA2-Uni \cite{ai2026llada2}  & 67.2 & 52.7 & 19.3 & 32.8 & 35.1 & 31.5 & 48.0 \\
LaViDa-O \cite{li2025lavidao}    & 54.0 & 36.1 & 13.8 & 27.0 & 23.5 & 34.2 & 34.5 \\

\rowcolor{cyan!10}
\ours-Base
& 71.5 & 68.3 & 35.8 & 35.5 & 54.4 & 52.4 & 59.8 \\
\rowcolor{cyan!10}
+ Naive OPD* & 47.7 & 46.7 & 17.9 & 33.4 & 23.0 & 38.7 & 42.0 \\

\rowcolor{cyan!10}
+ Cross-Paradigm OPD (\ours)
& 77.0 & 74.7 & 45.4 & 45.3 & 67.9 & 55.9 & 74.9 \\
\rowcolor{cyan!10}
\textit{$\Delta $ vs. Base }
& \textcolor{green!60!black}{\textit{+5.5}}
& \textcolor{green!60!black}{\textit{+6.4}}
& \textcolor{green!60!black}{\textit{+9.6}}
& \textcolor{green!60!black}{\textit{+9.8}}
& \textcolor{green!60!black}{\textit{+13.5}}
& \textcolor{green!60!black}{\textit{+3.5}}
& \textcolor{green!60!black}{\textit{+15.1}} \\

\hline

\end{tabular}
}
\end{table*}

\textbf{Text-to-Image}. We report results on GenEval \citep{ghosh2023geneval}, DPG \citep{hu2024equipdpg}, and HPS v3 \citep{ma2025hpsv3} for a comprehensive evaluation.  \ours~achieves strong results on these benchmarks and outperforms existing unified diffusion MLLMs such as MMaDA \citep{yang2025mmada} and LaViDa-O\citep{li2025lavidao}. Most notably, on HPSv3 which measures visual quality, \ours~closes the gap between discrete diffusion models and continuous diffusion models and even surpasses strong baselines like Flux.1-dev \citep{flux2024} and Qwen-Image \citep{wu2025qwen}, highlighting the effectiveness of the self-refining process.

\begin{table*}[t!]
  \centering
  \caption{\textbf{Text-to-Image Generation Performance on GenEval, DPG, and HPSv3.} $\dagger$ LLaDa-2-Uni incorporates a pretrained 6B continuous diffusion decoder. *Half of the understanding layers are not loaded in T2I tasks due to the Elastic-MoT design. }
  \label{tab:gen_eval_results}
\resizebox{\textwidth}{!}{
\scriptsize
\setlength{\tabcolsep}{3pt}
  \begin{tabular}{cccccccccccc}
    \toprule
    & & & \multicolumn{7}{c}{\textbf{GenEval}} & \textbf{DPG} & \textbf{HPSv3} \\
    \cmidrule(lr){4-10} \cmidrule(lr){11-11} \cmidrule(lr){12-12}
    \textbf{Model}  & \textbf{Family} & \textbf{Params} & \textbf{Single Obj.$\uparrow$} & \textbf{Two Obj.$\uparrow$} & \textbf{Counting$\uparrow$} & \textbf{Colors$\uparrow$} & \textbf{Position$\uparrow$} & \textbf{Color Attri.$\uparrow$} & \textbf{Overall$\uparrow$} & \textbf{Overall$\uparrow$} & \textbf{Score$\uparrow$} \\

        \midrule
  \multicolumn{12}{c}{\textit{Gen. Only}} \\
    SD3-Medium~\citep{esser2024scaling-sd3} & Cont. & 2B & 0.99 & 0.94 & 0.72 & 0.89 & 0.33 & 0.60 & 0.74 & 84.1 & - \\
    FLUX.1-dev~\citep{flux2024} & Cont. & 12B & 0.98 & 0.81 & 0.74 & 0.79 & 0.22 & 0.45 & 0.66 & 83.8 & 10.88 \\
    Qwen-Image\citep{wu2025qwen} & Cont. & 20B & 0.99 & 0.92&  0.89&  0.88&  0.76 & 0.77& 0.87 & 88.3 & 10.91 \\
        Meissonic~\citep{bai2024meissonic} & Disc. & 1B & 0.99 & 0.66 & 0.42 & 0.86 & 0.10 & 0.22 & 0.54 & - & - \\

   NLD-Image\citep{li2026nemotron} & Disc. & 8B & 0.98 & 0.93 & 0.83 & 0.94 & 0.88 & 0.82 & 0.90 & 85.2 & 9.61 \\
        \midrule
\multicolumn{12}{c}{\textit{Unified MLLM}} \\
        LLaDa-2-Uni \citep{ai2026llada2} & Hybrid & 16B+6B $\dagger$ & 1.00 & 0.98 & 0.73 & 0.92 & 0.90 & 0.84 & 0.89 & 87.8 & 9.47 \\
            BAGEL~\citep{deng2025emerging}  & Cont. & 14B & 0.98 & 0.95 & 0.84 & 0.95 & 0.78 & 0.77 & 0.88 & - & - \\
    Emu3~\citep{wang2024emu3} & Disc. & 8B & - & - & - & - & - & - & 0.66 & 80.6 & - \\
    Janus-Pro~\citep{chen2025janus}& Disc. & 7B & 0.99 & 0.89 & 0.59 & 0.90 & 0.79 & 0.66 & 0.80 & 84.1 & 8.81 \\
    MMaDA~\citep{yang2025mmada} & Disc.& 8B & 0.99 & 0.76 & 0.61 & 0.84 & 0.20 & 0.37 & 0.63 & 53.4 & 5.43 \\
    Show-o~\citep{xie2024show} & Disc. & 1.3B & 0.98 & 0.80 & 0.66 & 0.84 & 0.31 & 0.50 & 0.68 & - & 7.20 \\

    LaViDa-O~\citep{li2025lavidao} & Disc. & 10B & 0.99 & 0.85 & 0.71 & 0.86 & 0.65 & 0.58 & 0.77 & 81.8 & 8.81 \\
    \rowcolor{cyan!10}
    \ours  & Discrete. & 12B*  & 0.99 & 0.96 & 0.79 & 0.87 & 0.90 & 0.82 & 0.89 & 86.6 & 11.12 \\
    \bottomrule
  \end{tabular}
  \vspace{-20pt}
}
\end{table*}

\textbf{Image-Editing}. We report ImgEdit benchmark \citep{ye2025imgedit} in Table \ref{tab:image-edit}. \ours~outperforms discrete diffusion models like LaViDa-O \citep{li2025lavidao} and RL-post-trained LaViDa-R1\citep{li2026lavida}. Compared with a boarder range of model families, \ours~only underperforms GPT-4o and the Qwen-Image teacher, greatly reducing the gap between discrete diffusion models and the state-of-the-art works.  Compared with pre-OPD checkpoint \ours-Base, our cross-paradigm OPD leads to substantial improvements (+0.36). We also evaluate an off-policy KD baseline which supervises the student with independent teacher-generated samples. Our OPD framework, which uses teacher-refined student samples, outperforms the offline setup by a considerable margin (+0.27). We provide more discussion in the appendix.

\begin{table*}[t]
\centering
\caption{\textbf{Per-Category and overall scores on ImgEdit benchmark.} *Off-policy indicates we use an independent teacher output as the target, as opposed to teacher-refined student output.}
\label{tab:image-edit}
\resizebox{\textwidth}{!}{
\scriptsize
\setlength{\tabcolsep}{3pt}
\begin{tabular}{lcccccccccc}
\hline
\textbf{Model} & \textbf{Add} & \textbf{Adjust} & \textbf{Extract} & \textbf{Replace} & \textbf{Remove} & \textbf{Action} & \textbf{Background} & \textbf{Style} & \textbf{Hybrid} & \textbf{Overall} \\
\hline
\multicolumn{11}{c}{\textit{Other Methods}} \\
Qwen-Image \citep{wu2025qwen} & 4.38 & 4.16 & 3.43 & 4.66 & 4.14 & 4.69 & 4.38 & 4.81 & 3.82 & 4.27 \\
GPT-4o \citep{openai2024gpt4o} & 4.61 & 4.33 & 2.90 & 4.35 & 3.66 & 4.89 & 4.57 & 4.93 & 3.96 & 4.20 \\
LLaDa2-Uni \citep{ai2026llada2}& 3.76  &4.16  & 2.40  & 4.04  & 3.82  &4.42 & 4.07  & 4.60 & 3.97  & 3.92 \\
Qwen2.5VL+Flux \citep{wang2025gpt} & 4.07 & 3.79 & 2.04 & 4.13 & 3.89 & 4.52 & 3.90 & 4.84 & 3.04 & 3.80 \\

FluxKontext dev \citep{labs2025flux1kontextflowmatching} & 3.76 & 3.45 & 2.15 & 3.98 & 2.94 & 4.26 & 3.78 & 4.38 & 2.96 & 3.52 \\
OmniGen2 \citep{wu2025omnigen2}& 3.57 & 3.06 & 1.77 & 3.74 & 3.20 & 4.68 & 3.57 & 4.81 & 2.52 & 3.44 \\
BAGEL \citep{deng2025emerging}& 3.56 & 3.31 & 1.70 & 3.30 & 2.62 & 4.17 & 3.24 & 4.49 & 2.38 & 3.20 \\
Step1X-Edit \citep{liu2025step1x}  & 3.88 & 3.14 & 1.76 & 3.40 & 2.41 & 2.52 & 3.16 & 4.63 & 2.64 & 3.06 \\
InstructAny2Pix\citep{li2023instructany2pix} & 2.55 & 1.83 & 2.10 & 2.54 & 1.17 & 1.98 & 2.01 & 3.51 & 1.42 & 2.12 \\ 
Instruct-Pix2Pix\citep{brooks2023instructpix2pix} & 2.45 & 1.83 & 1.44 & 2.01 & 1.50 & 1.46 & 1.44 & 3.55 & 1.20 & 1.88 \\
\hline
\multicolumn{11}{c}{\textit{Discrete Diffusion Models}} \\
LaViDa-O \citep{li2025lavidao} & 4.04 & 3.62 & 2.01 & 4.39 & 3.98 & 3.54 & 4.06 & 4.82 & 2.94 & 3.71 \\

LaViDa-R1\citep{li2026lavida} & 4.25 & 3.90 & 2.32 & 4.52 & 4.06 & 4.18 & 3.86 & 4.87 & 3.10 & 3.90 \\
\rowcolor{cyan!10}
\ours-Base & 4.11 & 4.02 & 2.26 & 4.13 & 3.83 & 4.19 & 4.06 & 4.49 & 3.23 & 3.81 \\
\rowcolor{cyan!10}
+ Off-Policy KD* & 4.13 & 4.15 & 2.45 & 4.51 & 3.97 & 4.21 & 4.21 & 4.53 & 2.97 & 3.90 \\ 
\rowcolor{cyan!10}
+ Cross-Paradigm OPD (\ours) & 4.26 & 4.33 & 3.59 & 4.63 & 4.20 & 4.06 & 4.31 & 4.64 & 3.48 & 4.17 \\

\hline
\end{tabular}
}

\end{table*}

\subsection{Ablation Studies}

\begin{figure}
    \centering
    \includegraphics[width=1.0\linewidth]{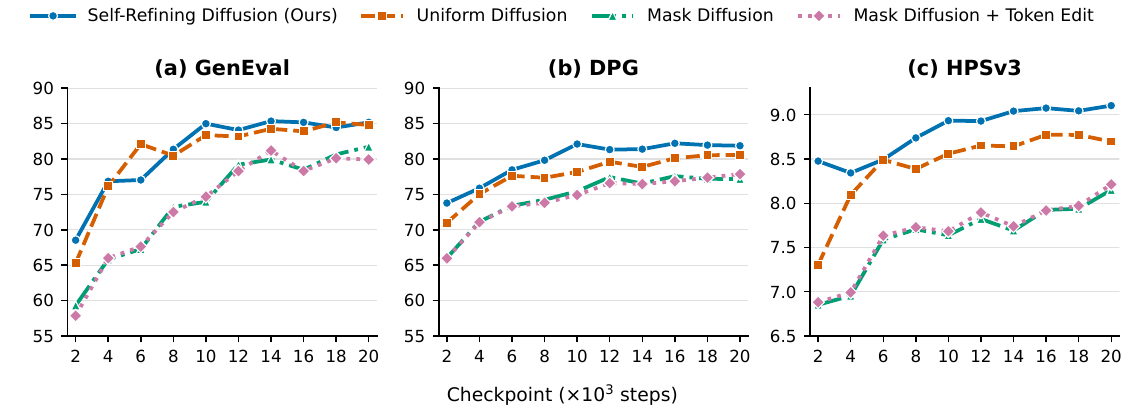}
    \caption{\textbf{Ablation Studies on the Self-Refining Diffusion Process.} We report performance on GenEval (a) DPG (b) and HPSv3 (c) benchmarks at different training steps using different paradigm.}
    \label{fig:sf_diff_appendix}
\end{figure}

\textbf{Self-Refining Diffusion Process.} We compare with the uniform diffusion, masked diffusion, and masked diffusion with token editing \citep{li2026nemotron} on GenEval, DPG, and HPSv3 in Figure \ref{fig:sf_diff_appendix}. We use 64 diffusion steps for all evaluations. We show the performance of models are trained with each method throughout a 20k-step run. On average, the proposed self-refining diffusion process performs the best. \ours~ outperforms masked diffusion overall. Compared with uniform diffusion,\ours~is comparable on GenEval, which checks high-level semantics using an object detector. On more fine-grained benchmarks like DPG and human-preference-aligned benchmarks like HPSv3, \ours consistently outperforms the uniform baseline. Among all benchmarks, the improvements on HPSv3 is the most significant, suggesting that the improvements of self-refinement is best for improving visual fidelity.

\textbf{Cross-Paradigm On-policy Distillation.} In Table \ref{tab:math-reasoning-results} and \ref{tab:image-edit}, we compare our proposed OPD framework with a naive OPD baseline on multimodal reasoning and an off-policy KD baseline on image editing tasks. We provide more experiment studies, such as the design of each loss term, the relation between AR performance and diffusion performance, and the effect of noise ratio in teacher refinement step to Appendix \ref{sec:appendix_ablation}.

\subsection{Qualitative Results}
\begin{figure}
    \centering
    \includegraphics[width=1.0\linewidth]{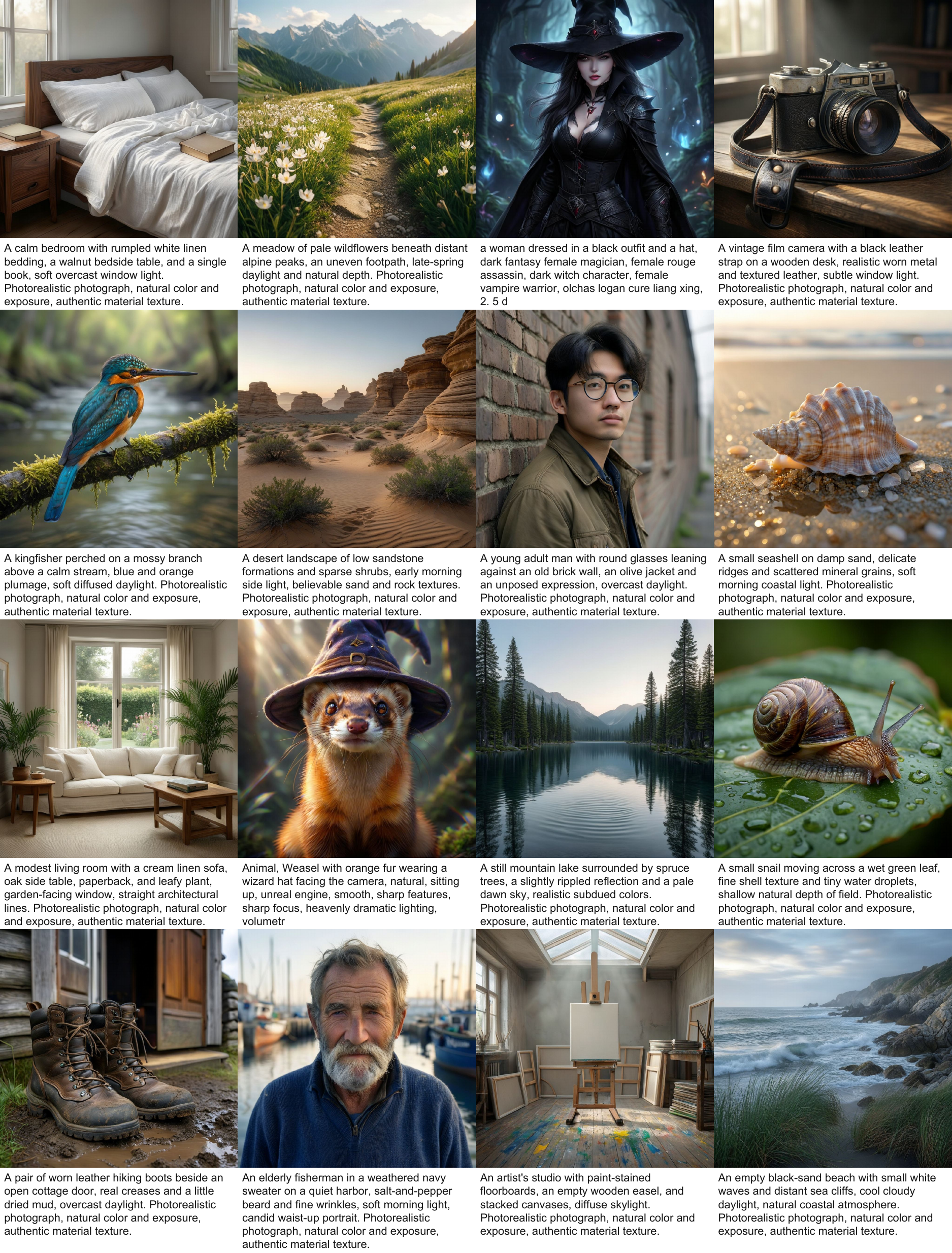}
    \caption{\textbf{Qualitative Results on Text-to-Image Tasks}. Results are sampled at $1024\times1024$ resolution. }
    \label{fig:appendix_t2idemo}
\end{figure}

\begin{figure}
    \centering
    \includegraphics[width=0.9\linewidth]{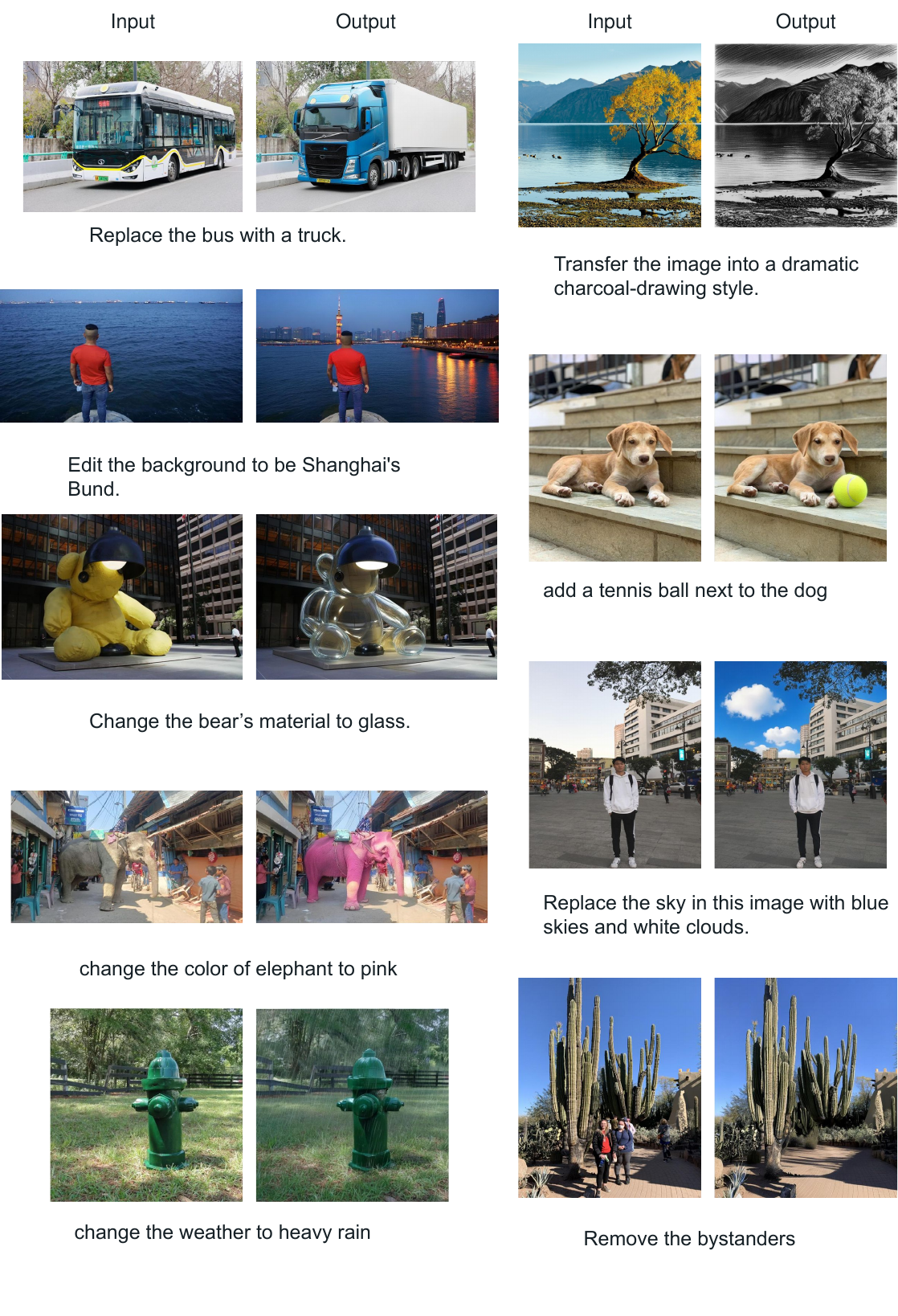}
    \caption{\textbf{Qualitative Results on Image Editing Tasks}. For each input-output pair, we present the input image on the left and the output image on the right. }
    \label{fig:appendix_editdemo}
\end{figure}

We present additional qualitative results of text-to-image generation in Figure \ref{fig:appendix_t2idemo} and image editing in Figure \ref{fig:appendix_editdemo}. These results highlight the strong generation capability of \ours.

\section{Conclusion}
Diffusion language models have attracted considerable attention in recent years for their greater efficiency compared with autoregressive models and their unified interface for multimodal data. However, their performance in practice still lags behind that of state-of-the-art models, limiting their adoption.
In this paper, we propose two methods for developing more capable DLLMs: cross-paradigm on-policy distillation, which transfers knowledge from diverse teacher models, and a self-refining diffusion process, which improves visual generation quality.
Extensive experiments demonstrate the effectiveness of both approaches. We hope \ours~will offer valuable insights for future research on DLLMs and unified multimodal systems.

{
  \small
  \bibliographystyle{unsrt}
  \bibliography{iclr2027_conference}

@inproceedings{agarwal2024gkd,
  title = {{On-Policy Distillation of Language Models: Learning from Self-Generated Mistakes}},
  author = {Agarwal, Rishabh and Vieillard, Nino and Zhou, Yongchao and Stanczyk, Piotr and Ramos, Sabela and Geist, Matthieu and Bachem, Olivier},
  booktitle = {International Conference on Learning Representations},
  year = {2024},
  url = {https://openreview.net/forum?id=3zKtaqxLhW},
}

@inproceedings{gu2024minillm,
  title = {{{MiniLLM}: Knowledge Distillation of Large Language Models}},
  author = {Gu, Yuxian and Dong, Li and Wei, Furu and Huang, Minlie},
  booktitle = {International Conference on Learning Representations},
  year = {2024},
  url = {https://openreview.net/forum?id=5h0qf7IBZZ},
}

@misc{niu2026tokenizer,
  title = {{Breaking the Tokenizer Barrier: On-Policy Distillation across Model Families}},
  author = {Niu, Yifan and Xiao, Han and Liu, Dongyi and Wang, Zelong and Gong, Dihong and Wang, Yasheng and Li, Jia},
  year = {2026},
  eprint = {2606.09456},
  archivePrefix = {arXiv},
  primaryClass = {cs.LG},
  url = {https://arxiv.org/abs/2606.09456},
}

@misc{su2026opdlm,
  title = {{Data-Efficient Autoregressive-to-Diffusion Language Models via On-Policy Distillation}},
  author = {Su, Xingyu and Helwig, Jacob and Parashar, Shubham and Chagi, Atharv and Jotsna, Lakshmi and Zhi, Degui and Caverlee, James and Kalathil, Dileep and Ji, Shuiwang},
  year = {2026},
  eprint = {2606.06712},
  archivePrefix = {arXiv},
  primaryClass = {cs.CL},
  url = {https://arxiv.org/abs/2606.06712},
}

@misc{ma2026mopd,
  title = {{{MOPD}: Multi-Teacher On-Policy Distillation for Capability Integration in {LLM} Post-Training}},
  author = {Ma, Wenhan and Wei, Jianyu and Zhao, Liang and Zhang, Hailin and Xiao, Bangjun and Li, Lei and Yang, Qibin and Gao, Bofei and Wang, Yudong and Li, Rang and Dong, Jinhao and Sui, Zhifang and Luo, Fuli},
  year = {2026},
  eprint = {2606.30406},
  archivePrefix = {arXiv},
  primaryClass = {cs.CL},
  url = {https://arxiv.org/abs/2606.30406},
}

@misc{li2026diffusionopd,
  title = {{{DiffusionOPD}: A Unified Perspective of On-Policy Distillation in Diffusion Models}},
  author = {Li, Quanhao and Yu, Junqiu and Jiang, Kaixun and Wei, Yujie and Xing, Zhen and Li, Pandeng and Chu, Ruihang and Zhang, Shiwei and Liu, Yu and Wu, Zuxuan},
  year = {2026},
  eprint = {2605.15055},
  archivePrefix = {arXiv},
  primaryClass = {cs.LG},
  url = {https://arxiv.org/abs/2605.15055},
}

@article{li2024llava,
  title = {{Llava-onevision: Easy visual task transfer}},
  author = {Li, Bo and Zhang, Yuanhan and Guo, Dong and Zhang, Renrui and Li, Feng and Zhang, Hao and Zhang, Kaichen and Zhang, Peiyuan and Li, Yanwei and Liu, Ziwei and Li, Chunyuan},
  journal = {arXiv preprint arXiv:2408.03326},
  year = {2024},
  eprint = {2408.03326},
  archivePrefix = {arXiv},
  url = {https://arxiv.org/abs/2408.03326},
}

@article{bai2025qwen25-vl,
  title = {{Qwen2.5-VL Technical Report}},
  author = {Bai, Shuai and Chen, Keqin and Liu, Xuejing and Wang, Jialin and Ge, Wenbin and Song, Sibo and Dang, Kai and Wang, Peng and Wang, Shijie and Tang, Jun and others},
  journal = {arXiv preprint arXiv:2502.13923},
  year = {2025},
  eprint = {2502.13923},
  archivePrefix = {arXiv},
  url = {https://arxiv.org/abs/2502.13923},
}

@article{openai2024gpt4o,
  title = {{GPT-4o System Card}},
  author = {OpenAI},
  journal = {arXiv preprint arXiv:2410.21276},
  year = {2024},
  eprint = {2410.21276},
  archivePrefix = {arXiv},
  url = {https://arxiv.org/abs/2410.21276},
}

@article{lou2023discrete-sedd,
  title = {{Discrete diffusion modeling by estimating the ratios of the data distribution}},
  author = {Lou, Aaron and Meng, Chenlin and Ermon, Stefano},
  journal = {arXiv preprint arXiv:2310.16834},
  year = {2023},
  eprint = {2310.16834},
  archivePrefix = {arXiv},
  url = {https://arxiv.org/abs/2310.16834},
}

@article{sahoo2024simple,
  title = {{Simple and effective masked diffusion language models}},
  author = {Sahoo, Subham and Arriola, Marianne and Schiff, Yair and Gokaslan, Aaron and Marroquin, Edgar and Chiu, Justin and Rush, Alexander and Kuleshov, Volodymyr},
  journal = {Advances in Neural Information Processing Systems},
  volume = {37},
  pages = {130136--130184},
  year = {2024},
}

@article{arriola2025block,
  title = {{Block diffusion: Interpolating between autoregressive and diffusion language models}},
  author = {Arriola, Marianne and Gokaslan, Aaron and Chiu, Justin T and Yang, Zhihan and Qi, Zhixuan and Han, Jiaqi and Sahoo, Subham Sekhar and Kuleshov, Volodymyr},
  journal = {arXiv preprint arXiv:2503.09573},
  year = {2025},
  eprint = {2503.09573},
  archivePrefix = {arXiv},
  url = {https://arxiv.org/abs/2503.09573},
}

@article{nie2025large,
  title = {{Large language diffusion models}},
  author = {Nie, Shen and Zhu, Fengqi and You, Zebin and Zhang, Xiaolu and Ou, Jingyang and Hu, Jun and Zhou, Jun and Lin, Yankai and Wen, Ji-Rong and Li, Chongxuan},
  journal = {arXiv preprint arXiv:2502.09992},
  year = {2025},
  eprint = {2502.09992},
  archivePrefix = {arXiv},
  url = {https://arxiv.org/abs/2502.09992},
}

@inproceedings{esser2024scaling-sd3,
  title = {{Scaling rectified flow transformers for high-resolution image synthesis}},
  author = {Esser, Patrick and Kulal, Sumith and Blattmann, Andreas and Entezari, Rahim and M{\"u}ller, Jonas and Saini, Harry and Levi, Yam and Lorenz, Dominik and Sauer, Axel and Boesel, Frederic and others},
  booktitle = {Forty-first international conference on machine learning},
  year = {2024},
}

@misc{chen2024open,
  title = {{Open-LLaVA-NeXT: An open-source implementation of LLaVA-NeXT series for facilitating the large multi-modal model community.}},
  author = {Chen, Lin and Xing, Long},
  year = {2024},
  doi = {10.5281/zenodo.13935471},
  url = {https://github.com/xiaoachen98/Open-LLaVA-NeXT},
}

@misc{liu2024llavanext,
  title = {{LLaVA-NeXT: Improved reasoning, OCR, and world knowledge}},
  author = {Liu, Haotian and Li, Chunyuan and Li, Yuheng and Li, Bo and Zhang, Yuanhan and Shen, Sheng and Lee, Yong Jae},
  year = {2024},
  month = {January},
  url = {https://llava-vl.github.io/blog/2024-01-30-llava-next/},
}

@inproceedings{yue2023mmmu,
  title = {{MMMU: A Massive Multi-discipline Multimodal Understanding and Reasoning Benchmark for Expert AGI}},
  author = {Xiang Yue and Yuansheng Ni and Kai Zhang and Tianyu Zheng and Ruoqi Liu and Ge Zhang and Samuel Stevens and Dongfu Jiang and Weiming Ren and Yuxuan Sun and Cong Wei and Botao Yu and Ruibin Yuan and Renliang Sun and Ming Yin and Boyuan Zheng and Zhenzhu Yang and Yibo Liu and Wenhao Huang and Huan Sun and Yu Su and Wenhu Chen},
  booktitle = {Proceedings of CVPR},
  year = {2024},
}

@article{lu2023mathvista,
  title = {{Mathvista: Evaluating mathematical reasoning of foundation models in visual contexts}},
  author = {Lu, Pan and Bansal, Hritik and Xia, Tony and Liu, Jiacheng and Li, Chunyuan and Hajishirzi, Hannaneh and Cheng, Hao and Chang, Kai-Wei and Galley, Michel and Gao, Jianfeng},
  journal = {arXiv preprint arXiv:2310.02255},
  year = {2023},
  eprint = {2310.02255},
  archivePrefix = {arXiv},
  url = {https://arxiv.org/abs/2310.02255},
}

@article{zhang2024mathverse,
  title = {{MathVerse: Does Your Multi-modal LLM Truly See the Diagrams in Visual Math Problems?}},
  author = {Zhang, Renrui and Jiang, Dongzhi and Zhang, Yichi and Lin, Haokun and Guo, Ziyu and Qiu, Pengshuo and Zhou, Aojun and Lu, Pan and Chang, Kai-Wei and Gao, Peng and others},
  journal = {arXiv preprint arXiv:2403.14624},
  year = {2024},
  eprint = {2403.14624},
  archivePrefix = {arXiv},
  url = {https://arxiv.org/abs/2403.14624},
}

@inproceedings{wang2024measuring,
  title = {{Measuring Multimodal Mathematical Reasoning with MATH-Vision Dataset}},
  author = {Ke Wang and Junting Pan and Weikang Shi and Zimu Lu and Houxing Ren and Aojun Zhou and Mingjie Zhan and Hongsheng Li},
  booktitle = {The Thirty-eighth Conference on Neural Information Processing Systems Datasets and Benchmarks Track},
  year = {2024},
  url = {https://openreview.net/forum?id=QWTCcxMpPA},
}

@inproceedings{lu2022learn,
  title = {{Learn to Explain: Multimodal Reasoning via Thought Chains for Science Question Answering}},
  author = {Lu, Pan and Mishra, Swaroop and Xia, Tony and Qiu, Liang and Chang, Kai-Wei and Zhu, Song-Chun and Tafjord, Oyvind and Clark, Peter and Ashwin Kalyan},
  booktitle = {The 36th Conference on Neural Information Processing Systems (NeurIPS)},
  year = {2022},
}

@inproceedings{kembhavi2016diagram,
  title = {{A diagram is worth a dozen images}},
  author = {Kembhavi, Aniruddha and Salvato, Mike and Kolve, Eric and Seo, Minjoon and Hajishirzi, Hannaneh and Farhadi, Ali},
  booktitle = {Computer Vision--ECCV 2016: 14th European Conference, Amsterdam, The Netherlands, October 11--14, 2016, Proceedings, Part IV 14},
  pages = {235--251},
  year = {2016},
  organization = {Springer},
}

@inproceedings{singh2019towards,
  title = {{Towards VQA Models That Can Read}},
  author = {Singh, Amanpreet and Natarajan, Vivek and Shah, Meet and Jiang, Yu and Chen, Xinlei and Batra, Dhruv and Parikh, Devi and Rohrbach, Marcus},
  booktitle = {Proceedings of the IEEE Conference on Computer Vision and Pattern Recognition},
  pages = {8317--8326},
  year = {2019},
}

@inproceedings{mathew2021docvqa,
  title = {{Docvqa: A dataset for vqa on document images}},
  author = {Mathew, Minesh and Karatzas, Dimosthenis and Jawahar, CV},
  booktitle = {Proceedings of the IEEE/CVF winter conference on applications of computer vision},
  pages = {2200--2209},
  year = {2021},
}

@inproceedings{masry-etal-2022-chartqa,
  title = {{{C}hart{QA}: A Benchmark for Question Answering about Charts with Visual and Logical Reasoning}},
  author = {Masry, Ahmed and Long, Do Xuan and Tan, Jia Qing and Joty, Shafiq and Hoque, Enamul},
  booktitle = {Findings of the Association for Computational Linguistics: ACL 2022},
  publisher = {Association for Computational Linguistics},
  address = {Dublin, Ireland},
  pages = {2263--2279},
  year = {2022},
  month = {May},
  doi = {10.18653/v1/2022.findings-acl.177},
  url = {https://aclanthology.org/2022.findings-acl.177},
}

@inproceedings{mathew2022infographicvqa,
  title = {{Infographicvqa}},
  author = {Mathew, Minesh and Bagal, Viraj and Tito, Rub{\`e}n and Karatzas, Dimosthenis and Valveny, Ernest and Jawahar, CV},
  booktitle = {Proceedings of the IEEE/CVF Winter Conference on Applications of Computer Vision},
  pages = {1697--1706},
  year = {2022},
}

@article{team2024gemini,
  title = {{Gemini 1.5: Unlocking multimodal understanding across millions of tokens of context}},
  author = {{Gemini Team} and Georgiev, Petko and Lei, Ving Ian and Burnell, Ryan and Bai, Libin and Gulati, Anmol and Tanzer, Garrett and Vincent, Damien and Pan, Zhufeng and Wang, Shibo and others},
  journal = {arXiv preprint arXiv:2403.05530},
  year = {2024},
  eprint = {2403.05530},
  archivePrefix = {arXiv},
  url = {https://arxiv.org/abs/2403.05530},
}

@inproceedings{chang2022maskgit,
  title = {{Maskgit: Masked generative image transformer}},
  author = {Chang, Huiwen and Zhang, Han and Jiang, Lu and Liu, Ce and Freeman, William T},
  booktitle = {Proceedings of the IEEE/CVF conference on computer vision and pattern recognition},
  pages = {11315--11325},
  year = {2022},
}

@article{yang2025mmada,
  title = {{MMaDA: Multimodal Large Diffusion Language Models}},
  author = {Yang, Ling and Tian, Ye and Li, Bowen and Zhang, Xinchen and Shen, Ke and Tong, Yunhai and Wang, Mengdi},
  journal = {arXiv preprint arXiv:2505.15809},
  year = {2025},
  eprint = {2505.15809},
  archivePrefix = {arXiv},
  url = {https://arxiv.org/abs/2505.15809},
}

@inproceedings{esser2021taming,
  title = {{Taming transformers for high-resolution image synthesis}},
  author = {Esser, Patrick and Rombach, Robin and Ommer, Bjorn},
  booktitle = {Proceedings of the IEEE/CVF conference on computer vision and pattern recognition},
  pages = {12873--12883},
  year = {2021},
}

@article{you2025lladav,
  title = {{Llada-v: Large language diffusion models with visual instruction tuning}},
  author = {You, Zebin and Nie, Shen and Zhang, Xiaolu and Hu, Jun and Zhou, Jun and Lu, Zhiwu and Wen, Ji-Rong and Li, Chongxuan},
  journal = {arXiv preprint arXiv:2505.16933},
  year = {2025},
  eprint = {2505.16933},
  archivePrefix = {arXiv},
  url = {https://arxiv.org/abs/2505.16933},
}

@article{li2025lavida,
  title = {{Lavida: A large diffusion language model for multimodal understanding}},
  author = {Li, Shufan and Kallidromitis, Konstantinos and Bansal, Hritik and Gokul, Akash and Kato, Yusuke and Kozuka, Kazuki and Kuen, Jason and Lin, Zhe and Chang, Kai-Wei and Grover, Aditya},
  journal = {arXiv preprint arXiv:2505.16839},
  year = {2025},
  eprint = {2505.16839},
  archivePrefix = {arXiv},
  url = {https://arxiv.org/abs/2505.16839},
}

@article{yu2025dimple,
  title = {{Dimple: Discrete diffusion multimodal large language model with parallel decoding}},
  author = {Yu, Runpeng and Ma, Xinyin and Wang, Xinchao},
  journal = {arXiv preprint arXiv:2505.16990},
  year = {2025},
  eprint = {2505.16990},
  archivePrefix = {arXiv},
  url = {https://arxiv.org/abs/2505.16990},
}

@article{chen2025blip3,
  title = {{Blip3-o: A family of fully open unified multimodal models-architecture, training and dataset}},
  author = {Chen, Jiuhai and Xu, Zhiyang and Pan, Xichen and Hu, Yushi and Qin, Can and Goldstein, Tom and Huang, Lifu and Zhou, Tianyi and Xie, Saining and Savarese, Silvio and others},
  journal = {arXiv preprint arXiv:2505.09568},
  year = {2025},
  eprint = {2505.09568},
  archivePrefix = {arXiv},
  url = {https://arxiv.org/abs/2505.09568},
}

@article{deng2025emerging,
  title = {{Emerging properties in unified multimodal pretraining}},
  author = {Deng, Chaorui and Zhu, Deyao and Li, Kunchang and Gou, Chenhui and Li, Feng and Wang, Zeyu and Zhong, Shu and Yu, Weihao and Nie, Xiaonan and Song, Ziang and others},
  journal = {arXiv preprint arXiv:2505.14683},
  year = {2025},
  eprint = {2505.14683},
  archivePrefix = {arXiv},
  url = {https://arxiv.org/abs/2505.14683},
}

@article{pan2025transfer,
  title = {{Transfer between modalities with metaqueries}},
  author = {Pan, Xichen and Shukla, Satya Narayan and Singh, Aashu and Zhao, Zhuokai and Mishra, Shlok Kumar and Wang, Jialiang and Xu, Zhiyang and Chen, Jiuhai and Li, Kunpeng and Juefei-Xu, Felix and others},
  journal = {arXiv preprint arXiv:2504.06256},
  year = {2025},
  eprint = {2504.06256},
  archivePrefix = {arXiv},
  url = {https://arxiv.org/abs/2504.06256},
}

@article{chen2025janus,
  title = {{Janus-pro: Unified multimodal understanding and generation with data and model scaling}},
  author = {Chen, Xiaokang and Wu, Zhiyu and Liu, Xingchao and Pan, Zizheng and Liu, Wen and Xie, Zhenda and Yu, Xingkai and Ruan, Chong},
  journal = {arXiv preprint arXiv:2501.17811},
  year = {2025},
  eprint = {2501.17811},
  archivePrefix = {arXiv},
  url = {https://arxiv.org/abs/2501.17811},
}

@article{liang2024mixture,
  title = {{Mixture-of-transformers: A sparse and scalable architecture for multi-modal foundation models}},
  author = {Liang, Weixin and Yu, Lili and Luo, Liang and Iyer, Srinivasan and Dong, Ning and Zhou, Chunting and Ghosh, Gargi and Lewis, Mike and Yih, Wen-tau and Zettlemoyer, Luke and others},
  journal = {arXiv preprint arXiv:2411.04996},
  year = {2024},
  eprint = {2411.04996},
  archivePrefix = {arXiv},
  url = {https://arxiv.org/abs/2411.04996},
}

@article{bai2024meissonic,
  title = {{Meissonic: Revitalizing masked generative transformers for efficient high-resolution text-to-image synthesis}},
  author = {Bai, Jinbin and Ye, Tian and Chow, Wei and Song, Enxin and Li, Xiangtai and Dong, Zhen and Zhu, Lei and Yan, Shuicheng},
  journal = {arXiv preprint arXiv:2410.08261},
  year = {2024},
  eprint = {2410.08261},
  archivePrefix = {arXiv},
  url = {https://arxiv.org/abs/2410.08261},
}

@article{besnier2025halton,
  title = {{Halton scheduler for masked generative image transformer}},
  author = {Besnier, Victor and Chen, Mickael and Hurych, David and Valle, Eduardo and Cord, Matthieu},
  journal = {arXiv preprint arXiv:2503.17076},
  year = {2025},
  eprint = {2503.17076},
  archivePrefix = {arXiv},
  url = {https://arxiv.org/abs/2503.17076},
}

@misc{flux2024,
  title = {{FLUX}},
  author = {{Black Forest Labs}},
  year = {2024},
  url = {https://github.com/black-forest-labs/flux},
}

@inproceedings{kazemzadeh2014referitgame,
  title = {{Referitgame: Referring to objects in photographs of natural scenes}},
  author = {Kazemzadeh, Sahar and Ordonez, Vicente and Matten, Mark and Berg, Tamara},
  booktitle = {Proceedings of the 2014 conference on empirical methods in natural language processing (EMNLP)},
  pages = {787--798},
  year = {2014},
}

@article{ghosh2023geneval,
  title = {{Geneval: An object-focused framework for evaluating text-to-image alignment}},
  author = {Ghosh, Dhruba and Hajishirzi, Hannaneh and Schmidt, Ludwig},
  journal = {Advances in Neural Information Processing Systems},
  volume = {36},
  pages = {52132--52152},
  year = {2023},
}

@article{ye2025imgedit,
  title = {{Imgedit: A unified image editing dataset and benchmark}},
  author = {Ye, Yang and He, Xianyi and Li, Zongjian and Lin, Bin and Yuan, Shenghai and Yan, Zhiyuan and Hou, Bohan and Yuan, Li},
  journal = {arXiv preprint arXiv:2505.20275},
  year = {2025},
  eprint = {2505.20275},
  archivePrefix = {arXiv},
  url = {https://arxiv.org/abs/2505.20275},
}

@article{lin2025uniworld,
  title = {{UniWorld: High-Resolution Semantic Encoders for Unified Visual Understanding and Generation}},
  author = {Lin, Bin and Li, Zongjian and Cheng, Xinhua and Niu, Yuwei and Ye, Yang and He, Xianyi and Yuan, Shenghai and Yu, Wangbo and Wang, Shaodong and Ge, Yunyang and others},
  journal = {arXiv preprint arXiv:2506.03147v1},
  year = {2025},
  eprint = {2506.03147v1},
  archivePrefix = {arXiv},
  url = {https://arxiv.org/abs/2506.03147v1},
  note = {Version 1},
}

@misc{labs2025flux1kontextflowmatching,
  title = {{FLUX.1 Kontext: Flow Matching for In-Context Image Generation and Editing in Latent Space}},
  author = {{Black Forest Labs} and Stephen Batifol and Andreas Blattmann and Frederic Boesel and Saksham Consul and Cyril Diagne and Tim Dockhorn and Jack English and Zion English and Patrick Esser and Sumith Kulal and Kyle Lacey and Yam Levi and Cheng Li and Dominik Lorenz and Jonas M\"uller and Dustin Podell and Robin Rombach and Harry Saini and Axel Sauer and Luke Smith},
  year = {2025},
  eprint = {2506.15742},
  archivePrefix = {arXiv},
  primaryClass = {cs.GR},
  url = {https://arxiv.org/abs/2506.15742},
}

@article{hu2024equipdpg,
  title = {{ELLA: Equip Diffusion Models with LLM for Enhanced Semantic Alignment}},
  author = {Hu, Xiwei and Wang, Rui and Fang, Yixiao and Fu, Bin and Cheng, Pei and Yu, Gang},
  journal = {arXiv preprint arXiv:2403.05135},
  year = {2024},
  eprint = {2403.05135},
  archivePrefix = {arXiv},
  url = {https://arxiv.org/abs/2403.05135},
}

@article{xie2024show,
  title = {{Show-o: One single transformer to unify multimodal understanding and generation}},
  author = {Xie, Jinheng and Mao, Weijia and Bai, Zechen and Zhang, David Junhao and Wang, Weihao and Lin, Kevin Qinghong and Gu, Yuchao and Chen, Zhijie and Yang, Zhenheng and Shou, Mike Zheng},
  journal = {arXiv preprint arXiv:2408.12528},
  year = {2024},
  eprint = {2408.12528},
  archivePrefix = {arXiv},
  url = {https://arxiv.org/abs/2408.12528},
}

@article{wu2025omnigen2,
  title = {{OmniGen2: Exploration to Advanced Multimodal Generation}},
  author = {Wu, Chenyuan and Zheng, Pengfei and Yan, Ruiran and Xiao, Shitao and Luo, Xin and Wang, Yueze and Li, Wanli and Jiang, Xiyan and Liu, Yexin and Zhou, Junjie and others},
  journal = {arXiv preprint arXiv:2506.18871v1},
  year = {2025},
  eprint = {2506.18871v1},
  archivePrefix = {arXiv},
  url = {https://arxiv.org/abs/2506.18871v1},
  note = {Version 1},
}

@article{wang2025gpt,
  title = {{GPT-IMAGE-EDIT-1.5M: A Million-Scale, GPT-Generated Image Dataset}},
  author = {Wang, Yuhan and Yang, Siwei and Zhao, Bingchen and Zhang, Letian and Liu, Qing and Zhou, Yuyin and Xie, Cihang},
  journal = {arXiv preprint arXiv:2507.21033},
  year = {2025},
  eprint = {2507.21033},
  archivePrefix = {arXiv},
  url = {https://arxiv.org/abs/2507.21033},
}

@inproceedings{brooks2023instructpix2pix,
  title = {{Instructpix2pix: Learning to follow image editing instructions}},
  author = {Brooks, Tim and Holynski, Aleksander and Efros, Alexei A},
  booktitle = {Proceedings of the IEEE/CVF conference on computer vision and pattern recognition},
  pages = {18392--18402},
  year = {2023},
}

@article{liu2025step1x,
  title = {{Step1x-edit: A practical framework for general image editing}},
  author = {Liu, Shiyu and Han, Yucheng and Xing, Peng and Yin, Fukun and Wang, Rui and Cheng, Wei and Liao, Jiaqi and Wang, Yingming and Fu, Honghao and Han, Chunrui and others},
  journal = {arXiv preprint arXiv:2504.17761},
  year = {2025},
  eprint = {2504.17761},
  archivePrefix = {arXiv},
  url = {https://arxiv.org/abs/2504.17761},
}

@misc{laion-aesthetics,
  title = {{LAION-AESTHETICS}},
  author = {Christoph Schuhmann},
  year = {2022},
  url = {https://laion.ai/blog/laion-aesthetics/},
  note = {Accessed: 2024 - 03 - 06},
}

@article{schuhmann2022laion,
  title = {{Laion-5b: An open large-scale dataset for training next generation image-text models}},
  author = {Schuhmann, Christoph and Beaumont, Romain and Vencu, Richard and Gordon, Cade and Wightman, Ross and Cherti, Mehdi and Coombes, Theo and Katta, Aarush and Mullis, Clayton and Wortsman, Mitchell and others},
  journal = {Advances in neural information processing systems},
  volume = {35},
  pages = {25278--25294},
  year = {2022},
}

@inproceedings{kirillov2023segment,
  title = {{Segment anything}},
  author = {Kirillov, Alexander and Mintun, Eric and Ravi, Nikhila and Mao, Hanzi and Rolland, Chloe and Gustafson, Laura and Xiao, Tete and Whitehead, Spencer and Berg, Alexander C and Lo, Wan-Yen and others},
  booktitle = {Proceedings of the IEEE/CVF international conference on computer vision},
  pages = {4015--4026},
  year = {2023},
}

@article{sun2023journeydb,
  title = {{Journeydb: A benchmark for generative image understanding}},
  author = {Sun, Keqiang and Pan, Junting and Ge, Yuying and Li, Hao and Duan, Haodong and Wu, Xiaoshi and Zhang, Renrui and Zhou, Aojun and Qin, Zipeng and Wang, Yi and others},
  journal = {Advances in neural information processing systems},
  volume = {36},
  pages = {49659--49678},
  year = {2023},
}

@article{chen2025sharegpt,
  title = {{ShareGPT-4o-Image: Aligning Multimodal Models with GPT-4o-Level Image Generation}},
  author = {Chen, Junying and Cai, Zhenyang and Chen, Pengcheng and Chen, Shunian and Ji, Ke and Wang, Xidong and Yang, Yunjin and Wang, Benyou},
  journal = {arXiv preprint arXiv:2506.18095},
  year = {2025},
  eprint = {2506.18095},
  archivePrefix = {arXiv},
  url = {https://arxiv.org/abs/2506.18095},
}

@article{guo2024mammoth,
  title = {{Mammoth-vl: Eliciting multimodal reasoning with instruction tuning at scale}},
  author = {Guo, Jarvis and Zheng, Tuney and Bai, Yuelin and Li, Bo and Wang, Yubo and Zhu, King and Li, Yizhi and Neubig, Graham and Chen, Wenhu and Yue, Xiang},
  journal = {arXiv preprint arXiv:2412.05237},
  year = {2024},
  eprint = {2412.05237},
  archivePrefix = {arXiv},
  url = {https://arxiv.org/abs/2412.05237},
}

@article{visualwebinstruct,
  title = {{VisualWebInstruct: Scaling up Multimodal Instruction Data through Web Search}},
  author = {Jia, Yiming and Li, Jiachen and Yue, Xiang and Li, Bo and Nie, Ping and Zou, Kai and Chen, Wenhu},
  journal = {arXiv preprint arXiv:2503.10582},
  year = {2025},
  eprint = {2503.10582},
  archivePrefix = {arXiv},
  url = {https://arxiv.org/abs/2503.10582},
}

@inproceedings{hanoona2023GLaMM,
  title = {{GLaMM: Pixel Grounding Large Multimodal Model}},
  author = {Rasheed, Hanoona and Maaz, Muhammad and Shaji, Sahal and Shaker, Abdelrahman and Khan, Salman and Cholakkal, Hisham and Anwer, Rao M. and Xing, Eric and Yang, Ming-Hsuan and Khan, Fahad S.},
  booktitle = {Proceedings of the IEEE/CVF Conference on Computer Vision and Pattern Recognition},
  pages = {13009--13018},
  year = {2024},
  url = {https://openaccess.thecvf.com/content/CVPR2024/html/Rasheed_GLaMM_Pixel_Grounding_Large_Multimodal_Model_CVPR_2024_paper.html},
}

@article{wu2025fast,
  title = {{Fast-dllm: Training-free acceleration of diffusion llm by enabling kv cache and parallel decoding}},
  author = {Wu, Chengyue and Zhang, Hao and Xue, Shuchen and Liu, Zhijian and Diao, Shizhe and Zhu, Ligeng and Luo, Ping and Han, Song and Xie, Enze},
  journal = {arXiv preprint arXiv:2505.22618},
  year = {2025},
  eprint = {2505.22618},
  archivePrefix = {arXiv},
  url = {https://arxiv.org/abs/2505.22618},
}

@article{li2023instructany2pix,
  title = {{Instructany2pix: Flexible visual editing via multimodal instruction following}},
  author = {Li, Shufan and Singh, Harkanwar and Grover, Aditya},
  journal = {arXiv preprint arXiv:2312.06738},
  year = {2023},
  eprint = {2312.06738},
  archivePrefix = {arXiv},
  url = {https://arxiv.org/abs/2312.06738},
}

@misc{kakaobrain2022coyo-700m,
  title = {{COYO-700M: Image-Text Pair Dataset}},
  author = {Byeon, Minwoo and Park, Beomhee and Kim, Haecheon and Lee, Sungjun and Baek, Woonhyuk and Kim, Saehoon},
  year = {2022},
  url = {https://github.com/kakaobrain/coyo-dataset},
}

@inproceedings{radford2021learning,
  title = {{Learning transferable visual models from natural language supervision}},
  author = {Radford, Alec and Kim, Jong Wook and Hallacy, Chris and Ramesh, Aditya and Goh, Gabriel and Agarwal, Sandhini and Sastry, Girish and Askell, Amanda and Mishkin, Pamela and Clark, Jack and others},
  booktitle = {International conference on machine learning},
  pages = {8748--8763},
  year = {2021},
  organization = {PMLR},
}

@article{mo2025x,
  title = {{X-fusion: Introducing new modality to frozen large language models}},
  author = {Mo, Sicheng and Nguyen, Thao and Huang, Xun and Iyer, Siddharth Srinivasan and Li, Yijun and Liu, Yuchen and Tandon, Abhishek and Shechtman, Eli and Singh, Krishna Kumar and Lee, Yong Jae and others},
  journal = {arXiv preprint arXiv:2504.20996},
  year = {2025},
  eprint = {2504.20996},
  archivePrefix = {arXiv},
  url = {https://arxiv.org/abs/2504.20996},
}

@article{shi2024lmfusion,
  title = {{LMFusion: Adapting Pretrained Language Models for Multimodal Generation}},
  author = {Shi, Weijia and Han, Xiaochuang and Zhou, Chunting and Liang, Weixin and Lin, Xi Victoria and Zettlemoyer, Luke and Yu, Lili},
  journal = {arXiv preprint arXiv:2412.15188},
  year = {2024},
  eprint = {2412.15188},
  archivePrefix = {arXiv},
  url = {https://arxiv.org/abs/2412.15188},
}

@inproceedings{shi2024simplified,
  title = {{Simplified and Generalized Masked Diffusion for Discrete Data}},
  author = {Shi, Jiaxin and Han, Kehang and Wang, Zhe and Doucet, Arnaud and Titsias, Michalis K.},
  booktitle = {Advances in Neural Information Processing Systems},
  year = {2024},
}

@article{li2025lavidao,
  title = {{Lavida-O: Elastic Masked Diffusion Models for Unified Multimodal Understanding and Generation}},
  author = {Li, Shufan and Gu, Jiuxiang and Liu, Kangning and Lin, Zhe and Wei, Zijun and Grover, Aditya and Kuen, Jason},
  journal = {arXiv preprint arXiv:2509.19244v1},
  year = {2025},
  eprint = {2509.19244v1},
  archivePrefix = {arXiv},
  url = {https://arxiv.org/abs/2509.19244v1},
  note = {Version 1},
}

@article{wang2024emu3,
  title = {{Emu3: Next-token prediction is all you need}},
  author = {Wang, Xinlong and Zhang, Xiaosong and Luo, Zhengxiong and Sun, Quan and Cui, Yufeng and Wang, Jinsheng and Zhang, Fan and Wang, Yueze and Li, Zhen and Yu, Qiying and others},
  journal = {arXiv preprint arXiv:2409.18869},
  year = {2024},
  eprint = {2409.18869},
  archivePrefix = {arXiv},
  url = {https://arxiv.org/abs/2409.18869},
}

@article{li2026nemotron,
  title = {{Nemotron-Labs-Diffusion-Image: Advancing Masked Discrete Diffusion for High-Resolution Image Synthesis}},
  author = {Li, Shufan and Heinrich, Greg and Ye, Hanrong and Fu, Yonggan and Grover, Aditya and Kautz, Jan and Molchanov, Pavlo},
  journal = {arXiv preprint arXiv:2606.29814},
  year = {2026},
  eprint = {2606.29814},
  archivePrefix = {arXiv},
  url = {https://arxiv.org/abs/2606.29814},
}

@article{liu2026visual,
  title = {{Visual-advantage on-policy distillation for vision-language models}},
  author = {Liu, Ruiqi and Lv, Xiaolei and Li, Gengsheng and Zhu, Ximo and Wang, Zhiheng and Zhang, Zhengbo and Chen, Junkai and Li, Zhiheng and Li, Bo and Gao, Jun and others},
  journal = {arXiv preprint arXiv:2605.21924},
  year = {2026},
  eprint = {2605.21924},
  archivePrefix = {arXiv},
  url = {https://arxiv.org/abs/2605.21924},
}

@article{wu2025qwen,
  title = {{Qwen-image technical report}},
  author = {Wu, Chenfei and Li, Jiahao and Zhou, Jingren and Lin, Junyang and Gao, Kaiyuan and Yan, Kun and Yin, Sheng-ming and Bai, Shuai and Xu, Xiao and Chen, Yilei and others},
  journal = {arXiv preprint arXiv:2508.02324},
  year = {2025},
  eprint = {2508.02324},
  archivePrefix = {arXiv},
  url = {https://arxiv.org/abs/2508.02324},
}

@article{gat2025set,
  title = {{Set block decoding is a language model inference accelerator}},
  author = {Gat, Itai and Ben-Hamu, Heli and Havasi, Marton and Haziza, Daniel and Reizenstein, Jeremy and Synnaeve, Gabriel and Lopez-Paz, David and Karrer, Brian and Lipman, Yaron},
  journal = {arXiv preprint arXiv:2509.04185},
  year = {2025},
  eprint = {2509.04185},
  archivePrefix = {arXiv},
  url = {https://arxiv.org/abs/2509.04185},
}

@inproceedings{cheng2026sdar,
  title = {{Sdar: A synergistic diffusion-autoregression paradigm for scalable sequence generation}},
  author = {Cheng, Shuang and Bian, Yihan and Liu, Dawei and Jiang, Yuhua and Liu, Yihao and Zhang, Linfeng and Yao, Qian and Tian, Zhongbo and Wang, Wenhai and Guo, Qipeng and others},
  booktitle = {Findings of the Association for Computational Linguistics: ACL 2026},
  pages = {22058--22075},
  year = {2026},
}

@article{zhu2026flare,
  title = {{FLARE: Diffusion for Hybrid Language Model}},
  author = {Zhu, Yuchen and Shi, Jing and Ge, Chongjian and Tan, Hao and Xu, Yiran and Zhu, Wanrong and Kuen, Jason and Goswami, Koustava and Jain, Rajiv and Chen, Yongxin and others},
  journal = {arXiv preprint arXiv:2606.01774},
  year = {2026},
  eprint = {2606.01774},
  archivePrefix = {arXiv},
  url = {https://arxiv.org/abs/2606.01774},
}

@article{wang2026remasking,
  title = {{Remasking discrete diffusion models with inference-time scaling}},
  author = {Wang, Guanghan and Schiff, Yair and Sahoo, Subham and Kuleshov, Volodymyr},
  journal = {Advances in Neural Information Processing Systems},
  volume = {38},
  pages = {147282--147339},
  year = {2025},
}

@article{chao2026beyond,
  title = {{Beyond masked and unmasked: Discrete diffusion models via partial masking}},
  author = {Chao, Chen-Hao Lance and Sun, Wei-Fang and Liang, Hanwen and Lee, Chun-Yi and Krishnan, Rahul},
  journal = {Advances in Neural Information Processing Systems},
  volume = {38},
  pages = {97458--97500},
  year = {2025},
}

@techreport{krizhevsky2009learning,
  title = {{Learning multiple layers of features from tiny images}},
  author = {Krizhevsky, Alex},
  institution = {University of Toronto},
  address = {Toronto, Ontario, Canada},
  year = {2009},
  url = {https://www.cs.toronto.edu/~kriz/learning-features-2009-TR.pdf},
}

@inproceedings{deng2009imagenet,
  title = {{Imagenet: A large-scale hierarchical image database}},
  author = {Deng, Jia and Dong, Wei and Socher, Richard and Li, Li-Jia and Li, Kai and Fei-Fei, Li},
  booktitle = {2009 IEEE conference on computer vision and pattern recognition},
  pages = {248--255},
  year = {2009},
  organization = {Ieee},
}

@article{hao2026stabilizing,
  title = {{Stabilizing On-Policy Distillation for MLLM Reasoning with Global Normalization}},
  author = {Hao, Dongze and Jin, Zhiwei and Chen, Chen and Lu, Haonan},
  journal = {arXiv preprint arXiv:2606.09091},
  year = {2026},
  eprint = {2606.09091},
  archivePrefix = {arXiv},
  url = {https://arxiv.org/abs/2606.09091},
}

@article{yuan2026vision,
  title = {{Vision-opd: Learning to see fine details for multimodal llms via on-policy self-distillation}},
  author = {Yuan, Qianhao and Lou, Jie and Yu, Xing and Lin, Hongyu and Sun, Le and Han, Xianpei and Lu, Yaojie},
  journal = {arXiv preprint arXiv:2605.18740},
  year = {2026},
  eprint = {2605.18740},
  archivePrefix = {arXiv},
  url = {https://arxiv.org/abs/2605.18740},
}

@article{fu2026nemotron,
  title = {{Nemotron-Labs-Diffusion: A Tri-Mode Language Model Unifying Autoregressive, Diffusion, and Self-Speculation Decoding}},
  author = {Fu, Yonggan and Whalen, Lexington and Garg, Abhinav and Wu, Chengyue and Khadkevich, Maksim and Oswald, Nicolai and Xie, Enze and Egert, Daniel and Sreenivas, Sharath Turuvekere and Diao, Shizhe and others},
  journal = {arXiv preprint arXiv:2607.05722},
  year = {2026},
  eprint = {2607.05722},
  archivePrefix = {arXiv},
  url = {https://arxiv.org/abs/2607.05722},
}

@inproceedings{heinrich2025radiov2,
  title = {{RADIOv2.5: Improved Baselines for Agglomerative Vision Foundation Models}},
  author = {Heinrich, Greg and Ranzinger, Mike and Yin, Hongxu and Lu, Yao and Kautz, Jan and Tao, Andrew and Catanzaro, Bryan and Molchanov, Pavlo},
  booktitle = {Proceedings of the IEEE/CVF Conference on Computer Vision and Pattern Recognition},
  pages = {22487--22497},
  year = {2025},
}

@inproceedings{huang2023lightweight,
  title = {{Lightweight-yet-efficient: Revitalizing ball-tree for point-to-hyperplane nearest neighbor search}},
  author = {Huang, Qiang and Tung, Anthony KH},
  booktitle = {2023 IEEE 39th International Conference on Data Engineering (ICDE)},
  pages = {436--449},
  year = {2023},
  organization = {IEEE},
}

@article{chen2024fast,
  title = {{Fast sampling via discrete non-markov diffusion models with predetermined transition time}},
  author = {Chen, Zixiang and Yuan, Huizhuo and Li, Yongqian and Kou, Yiwen and Zhang, Junkai and Gu, Quanquan},
  journal = {Advances in Neural Information Processing Systems},
  volume = {37},
  pages = {106870--106905},
  year = {2024},
}

@article{deshmukh2026nemotron,
  title = {{Nemotron 3 nano omni: efficient and open multimodal intelligence}},
  author = {{NVIDIA} and Deshmukh, Amala Sanjay and Chumachenko, Kateryna and Rintamaki, Tuomas and Le, Matthieu and Poon, Tyler and Taheri, Danial Mohseni and Karmanov, Ilia and Liu, Guilin and Seppanen, Jarno and Goel, Arushi and others},
  journal = {arXiv preprint arXiv:2604.24954},
  year = {2026},
  eprint = {2604.24954},
  archivePrefix = {arXiv},
  url = {https://arxiv.org/abs/2604.24954},
}

@misc{xai2024realworldqa,
  title = {{RealWorldQA Dataset}},
  author = {{xAI}},
  publisher = {Hugging Face},
  year = {2024},
  url = {https://huggingface.co/datasets/xai-org/RealworldQA},
}

@inproceedings{li2025screenspot,
  title = {{Screenspot-pro: Gui grounding for professional high-resolution computer use}},
  author = {Li, Kaixin and Meng, Ziyang and Lin, Hongzhan and Luo, Ziyang and Tian, Yuchen and Ma, Jing and Huang, Zhiyong and Chua, Tat-Seng},
  booktitle = {Proceedings of the 33rd ACM International Conference on Multimedia},
  pages = {8778--8786},
  year = {2025},
}

@inproceedings{qiao2025we,
  title = {{We-math: Does your large multimodal model achieve human-like mathematical reasoning?}},
  author = {Qiao, Runqi and Tan, Qiuna and Dong, Guanting and Wu, Minhui and Sun, Chong and Song, Xiaoshuai and Wang, Jiapeng and Gongque, Zhuoma and Lei, Shanglin and Zhang, Yifan and others},
  booktitle = {Proceedings of the 63rd Annual Meeting of the Association for Computational Linguistics (Volume 1: Long Papers)},
  pages = {20023--20070},
  year = {2025},
}

@inproceedings{lu2021inter,
  title = {{Inter-gps: Interpretable geometry problem solving with formal language and symbolic reasoning}},
  author = {Lu, Pan and Gong, Ran and Jiang, Shibiao and Qiu, Liang and Huang, Siyuan and Liang, Xiaodan and Zhu, Song-Chun},
  booktitle = {Proceedings of the 59th Annual Meeting of the Association for Computational Linguistics and the 11th International Joint Conference on Natural Language Processing (Volume 1: Long Papers)},
  pages = {6774--6786},
  year = {2021},
}

@inproceedings{ma2025hpsv3,
  title = {{Hpsv3: Towards wide-spectrum human preference score}},
  author = {Ma, Yuhang and Wu, Xiaoshi and Sun, Keqiang and Li, Hongsheng},
  booktitle = {2025 IEEE/CVF International Conference on Computer Vision (ICCV)},
  pages = {15086--15095},
  year = {2025},
  organization = {IEEE},
}

@article{li2026lavida,
  title = {{LaViDa-R1: Advancing reasoning for unified multimodal diffusion language models}},
  author = {Li, Shufan and Zhu, Yuchen and Gu, Jiuxiang and Liu, Kangning and Lin, Zhe and Chen, Yongxin and Tao, Molei and Grover, Aditya and Kuen, Jason},
  journal = {arXiv preprint arXiv:2602.14147},
  year = {2026},
  eprint = {2602.14147},
  archivePrefix = {arXiv},
  url = {https://arxiv.org/abs/2602.14147},
}

@article{jiang2026trajectory,
  title={Trajectory-Refined Distillation},
  author={Jiang, Li and Xu, Haoran and Ding, Yichuan and Zhang, Amy},
  journal={arXiv preprint arXiv:2606.08432},
  year={2026}
}

@article{bai2025qwen3,
  title={Qwen3-vl technical report},
  author={Bai, Shuai and Cai, Yuxuan and Chen, Ruizhe and Chen, Keqin and Chen, Xionghui and Cheng, Zesen and Deng, Lianghao and Ding, Wei and Gao, Chang and Ge, Chunjiang and others},
  journal={arXiv preprint arXiv:2511.21631},
  year={2025}
}

@article{chen2026scaleedit,
  title={ScaleEdit-12M: Scaling Open-Source Image Editing Data Generation via Multi-Agent Framework},
  author={Chen, Guanzhou and Cui, Erfei and Tian, Changyao and Yang, Danni and Yang, Ganlin and Qiao, Yu and Li, Hongsheng and Luo, Gen and Zhang, Hongjie},
  journal={arXiv preprint arXiv:2603.20644},
  year={2026}
}

@article{cui2025emu3,
  title={Emu3. 5: Native multimodal models are world learners},
  author={Cui, Yufeng and Chen, Honghao and Deng, Haoge and Huang, Xu and Li, Xinghang and Liu, Jirong and Liu, Yang and Luo, Zhuoyan and Wang, Jinsheng and Wang, Wenxuan and others},
  journal={arXiv preprint arXiv:2510.26583},
  year={2025}
}

@article{liu2026ministral,
  title={Ministral 3},
  author={Liu, Alexander H and Khandelwal, Kartik and Subramanian, Sandeep and Jouault, Victor and Rastogi, Abhinav and Sad{\'e}, Adrien and Jeffares, Alan and Jiang, Albert and Cahill, Alexandre and Gavaudan, Alexandre and others},
  journal={arXiv preprint arXiv:2601.08584},
  year={2026}
}

@article{nvidia2026cosmos3,
  title = {Cosmos 3: Omnimodal World Models for Physical AI},
  author = {{NVIDIA}},
  journal = {arXiv preprint arXiv:2606.02800},
  year = {2026},
  url = {https://research.nvidia.com/labs/cosmos-lab/cosmos3/technical-report.pdf}
}

@article{ai2026llada2,
  title={Llada2. 0-uni: Unifying multimodal understanding and generation with diffusion large language model},
  author={AI, Inclusion and Bie, Tiwei and Chen, Haoxing and Chen, Tieyuan and Cheng, Zhenglin and Cui, Long and Gan, Kai and Huang, Zhicheng and Lan, Zhenzhong and Li, Haoquan and others},
  journal={arXiv preprint arXiv:2604.20796},
  year={2026}
}

@article{wu2026fast,
  title={Fast-dVLM: Efficient Block-Diffusion VLM via Direct Conversion from Autoregressive VLM},
  author={Wu, Chengyue and Lan, Shiyi and Fu, Yonggan and Gao, Sensen and Wang, Jin and Yu, Jincheng and Alvarez, Jose M and Molchanov, Pavlo and Luo, Ping and Han, Song and others},
  journal={arXiv preprint arXiv:2604.06832},
  year={2026}
}

@article{lu2022unified,
  title={Unified-io: A unified model for vision, language, and multi-modal tasks},
  author={Lu, Jiasen and Clark, Christopher and Zellers, Rowan and Mottaghi, Roozbeh and Kembhavi, Aniruddha},
  journal={arXiv preprint arXiv:2206.08916},
  year={2022}
}

@article{fu2026ocrbench,
  title={Ocrbench v2: An improved benchmark for evaluating large multimodal models on visual text localization and reasoning},
  author={Fu, Ling and Kuang, Zhebin and Song, Jiajun and Huang, Mingxin and Yang, Biao and Li, Yuzhe and Zhu, Linghao and Luo, Qidi and Wang, Xinyu and Lu, Hao and others},
  journal={Advances in Neural Information Processing Systems},
  volume={38},
  year={2026}
}

@inproceedings{zou2025dynamath,
  title={Dynamath: A dynamic visual benchmark for evaluating mathematical reasoning robustness of vision language models},
  author={Zou, Chengke and Guo, Xingang and Yang, Rui and Zhang, Junyu and Hu, Bin and Zhang, Huan},
  booktitle={International Conference on Learning Representations},
  volume={2025},
  pages={48337--48383},
  year={2025}
}

@article{xiao2024logicvista,
  title={Logicvista: Multimodal llm logical reasoning benchmark in visual contexts},
  author={Xiao, Yijia and Sun, Edward and Liu, Tianyu and Wang, Wei},
  journal={arXiv preprint arXiv:2407.04973},
  year={2024}
}

@misc{moonshot_kimi_k26, title = {Kimi K2.6}, 
 author = {Moonshot AI},
 howpublished = {\url{https://huggingface.co/moonshotai/Kimi-K2.6}}, year = {2026} 
 }
}

\appendix
\section{Additional Technical Details}

\subsection{Formulations of SBD Models}
\label{sec:appendix_sbd}
\newcommand{\cat}[0]{\text{Cat}}
\newcommand{\alphats}[0]{\frac{1-t}{1-s}}
\newcommand{\oneminusalphats}[0]{\frac{t-s}{1-s}}

This section briefly describes the formulation of SBD models. \ours~largely adopts its canonical form.

Given a sequence of discrete tokens $X^0=[X_1^0,X_2^0,..X_L^0]$, AR models generate tokens by repetitively sampling from the next-token prediction $p(X_i^0|X^0_{j<i})$. By contrast, masked discrete diffusion models (MDMs) initialize a fully masked sequence $X^1=[[m],[m],...]$ and gradually unmask tokens during the continuous interval $[0,1]$ through the reverse process $p(X^s|X^t)$, where $0\le s \le t \le 1$, until the sequence reaches a clean state $X^0$ that consists of only non-mask tokens. At each step, tokens are sampled independently by assuming the factorization $p(X^s|X^t)=\prod_i p (X^s_i|X^t)$. Since $p (X^s_i|X^t)$ depends on the entire sequence $X^t$ for every token position $i$, MDMs employ bidirectional attention over the full sequence as opposed to causal attention in AR models. Consequently, MDMs suffer from efficiency bottlenecks despite being capable of generating multiple tokens in parallel.

To address this limitation, models like Block Diffusion \citep{arriola2025block}, SDAR \citep{cheng2026sdar} and SBD \citep{gat2025set} adopt a semi-autoregressive generation paradigm. 

Specifically, the full sequence $X^0=[X_1^0,X_2^0,..X_L^0]$ is broken down into multiple consecutive multi-token blocks  $X^0=[B_1^0,...B_K^0]$. The tokens within each block are generated with the standard MDM sampling process, but the blocks are sampled sequentially from left to right. This design only requires within-block bidirectional attention, greatly reducing the compute cost of fully bidirectional MDMs. 

Formally, let $X_{b,i}^t$ denote the $i$-th token in the $b$-th block. The forward diffusion process is defined as
\begin{equation}
    q(X_{b,i}^t|X_{b,i}^s) =  
    \begin{cases}
      \textbf{M}, & \text{if } X_{b,i}^s=[M] \\
      \alphats \mathbf{X_{b,i}^s}+\oneminusalphats \textbf{M}, & \text{if } X_{b,i}^s \ne [M],
    \end{cases}
    \label{eq:sbd_forward_transit}
\end{equation}
where $0\le s \le t \le 1$, and $\textbf{M},\mathbf{X_{b,i}^s}$ are one-hot probability vectors corresponding to the mask token \texttt{[m]} and the clean token $X_{b,i}^s$.

When $s=0$, we obtain the following marginal:
\begin{equation}
    q(X_{b,i}^t|X_{b,i}^0) = (1-t)\mathbf{X_{b,i}^0}+ t \textbf{M}.
        \label{eq:sbd_forward}
\end{equation}

The posterior of the reverse process $p(X^s|X^t,X^0)$ has the following form \citep{sahoo2024simple}:
\begin{equation}
    p(  X_{b,i}^s|X_{b,i}^0, X_{b,i}^t) =  
    \begin{cases}
   \mathbf{X_{b,i}^t} & \text{if } X_{b,i}^t \ne [M] \\
      \tfrac{t-s}{t} \mathbf{X_{b,i}^0}+\tfrac{s}{t} \textbf{M}, & \text{if } X_{b,i}^t = [M].
    \end{cases}
    \label{eq:appendix-eq-p}
\end{equation}.

Practically, a neural network $\theta$ is used to model the clean token prediction $p_\theta(X_{b,i}^0|B^0_{j<b},B_b^t)$ conditioned on the partially masked current block $B_b^t$ and all previous clean blocks $B^0_{j<b}$, which gives the following transition:
\begin{equation}
    p_\theta(  X_{b,i}^s|B_{j<b}^0, B_{b}^t) =  
    \begin{cases}
   \mathbf{X_{b,i}^t} & \text{if } X_{b,i}^t \ne [M] \\
      \tfrac{t-s}{t}p_\theta(X_{b,i}^0|B^0_{j<b},B_b^t)+\tfrac{s}{t} \textbf{M}, & \text{if } X_{b,i}^t = [M].
    \end{cases}
    \label{eq:appendix-inference}
\end{equation}. 

\begin{figure}[t]
    \centering
    \includegraphics[width=1.0\linewidth]{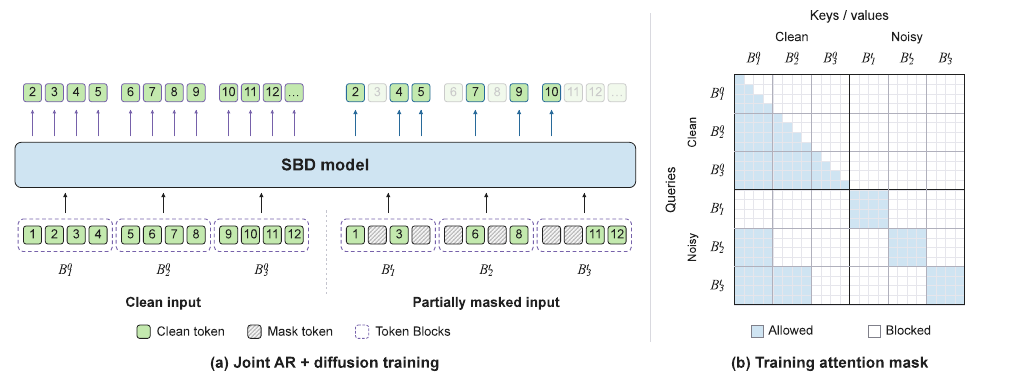}
    \caption{\textbf{The design of SBD formulation.} (Left) SBD adopts a two-stream training process that concatenates clean blocks and noisy blocks in the sequence dimension. It applies AR CE loss on the clean blocks, and diffusion CE loss on the noisy blocks. (Right) SBD models use a specialized attention mask with a causal mask in the clean blocks and block-bidirectional mask in the noisy blocks. Each noisy block can attend to all other tokens in itself and all tokens in preceding clean blocks. }
    \label{fig:sbd_training}
\end{figure}
\textbf{Training Process.} SBD employs a specialized attention mask to allow parallel training over multiple blocks. Given a sequence  $X^0=[X_1^0,X_2^0,..X_N^0]$ sampled from some data distribution $\mathcal{D}$ and a randomly sampled timestep $t\sim U([0,1])$, we first divide $X^0$ into consecutive blocks $B_1^0,...B_K^0$ and construct corresponding partially masked blocks $B_1^t,...B_K^t$ using the forward marginal in Equation \ref{eq:sbd_forward}. The input to the model is constructed by concatenating the clean sequence and partially masked sequence, obtaining $[B_1^0,...B_K^0,B_1^t,...B_K^t]$. A specialized attention mask is used to ensure the correct semantics and allow for parallel multi-block training. In particular, tokens in the clean block $B_i^0$ can attend to preceding clean tokens in $B_i^0$ and all tokens in preceding clean blocks $B_{j<i}^0$. Tokens in the partially masked block $B_i^t$ can attend to all tokens in the same partially masked block $B_i^t$ and all tokens in preceding clean blocks $B_{j<i}^0$. The loss target is simply the concatenated clean sequence $Y=X^0|X^0$ shifted by 1 position to the left. This design is illustrated in Figure \ref{fig:sbd_training}. 

At each training step, the block size is randomized. The loss is also applied on the causal clean token portion of the input, which corresponds to a block size of one and is effectively an AR view. We note that \ours~adopts a position-shifted variant of SBD \citep{zhu2026flare} to better align the AR-view and diffusion view with a block size larger than 1, as opposed to the vanilla version \cite{gat2025set} which does not shift tokens on the partially masked portion of the input and has a mismatch between the AR view and the block-size-one diffusion view.

In pure language modeling, a sequence is divided into equally sized blocks. \ours~further allows varying-sized blocks in the same sequence because it ensures tokens from the same image belong to the same block and can attend to each other.

\textbf{Sampling Process.} At inference, we sample blocks $B_1,B_2..$ sequentially until some block contains a termination token. Each block is initialized with a fully masked block $B_i^1$ and progressively unmasked via the parameterized reverse process in Equation \ref{eq:appendix-inference}. In particular, we discretize the time interval [0,1] into discrete steps $0=t_0<t_1<t_2..t_T=1$, and recursively query Equation \ref{eq:appendix-inference} starting from $B_i^1$ until we reach the clean state $B_i^0$.

\subsection{Formulations of Self-Refining Diffusion Process}

\begin{figure}
    \centering
    \includegraphics[width=1.0\linewidth]{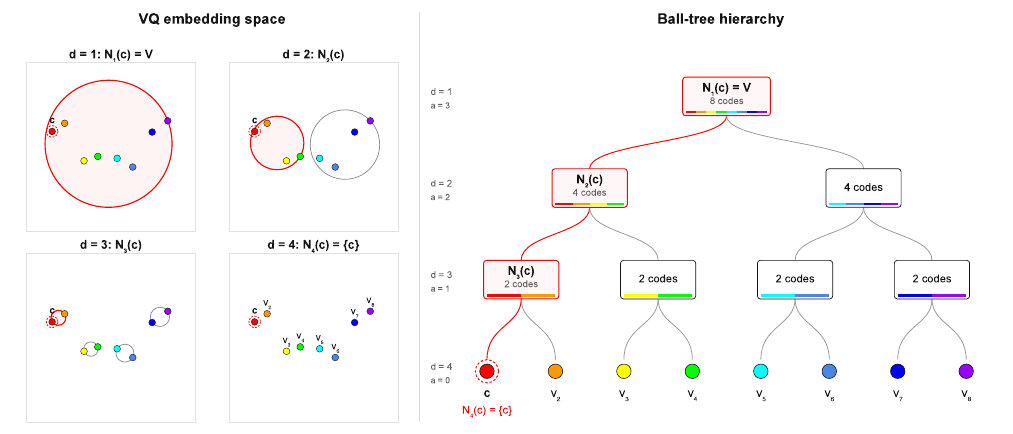}
    \caption{\textbf{Visualization of Ball-Tree-Based Corruption Process.} (Right) Ball-Tree is a balanced binary tree whose leaf nodes correspond to unique tokens in the vocabulary. The forward corruption process $U_{k|k-1}$ increases the token age $a$ and decreases the node depth $d$. (Left) Each node has a geometric representation which is a high-dimensional ball in the VQ embedding space. Decreasing depth $d$ corresponds to increasing ball radius. In this visualization, the depth $D=4$ and maximum token age $K=3$.   }
    \label{fig:appendix_corruption}
\end{figure}

For text blocks, we employ the standard MDM formulation described in Section \ref{sec:appendix_sbd} to minimize the gap with the base LLM NLD used to build \ours. For image blocks, we employ our proposed self-refining diffusion process with intermediate mask states. For reduced complexity and sampling efficiency, we adopt the non-Markov formulation (DNDM) \citep{chen2024fast}. For simplicity of notation, this discussion focuses on the setup with a single image sequence  $X^0=[X_1^0,...X_L^0]$ whose tokens belong to one block. However, it can be easily extended to multi-block settings with interleaved multi-modal sequences.

\subsubsection{Ball-Tree-Based Per-Token Corruption Process} 
\label{sec:appendix_ball_tree}

Recall from Section \ref{sec:self_refinement} that we adopt the state space $\mathcal{V} \times \mathcal{A}$ where $\mathcal{V}$ is the vocabulary of a tokenizer and $\mathcal{A}=\{0,1,...K\}$ represents the set of corruption levels that connect the fully masked state and the clean state. Each token $X_i^t$ in this state space is denoted as $X_i^t=(C_i^t,A_i^t)$ where $C_i^t\in \mathcal{V}$, $A_i^t \in \mathcal {A}$. In particular, $A_i=0$ represents a clean state and  $A_i=K$ represents a mask state. We define the age-dependent corruption process such that $A_i$ monotonically increases at state transitions in the forward process. 

Concretely, each token that starts at state $(c,a)$ for some $c\in\mathcal{V}, a \in \mathcal{A}$ can only transition to $(c',a+1)$, with $(*,0)$ being the set of initial states and $(*,K)$ being the set of absorbing mask states.  The corruption process can be expressed by the transition probability $U_{j|k}(u|c)$ for $0 \le k < j \le K$. It must satisfy the constraint: 
\begin{align}
    U_{K|k}(u|c)= Q(u),
\end{align}
which implies that the absorbing state $Q$ is an uninformative stationary distribution over possible mask states that is not dependent on the initial clean token $c$. Ideally we want the transition to be progressive so that the state becomes gradually less informative about the clean token $c$ when we progress from $a=0$ to $a=K$.

We implement this as a ball tree-based corruption process. A ball tree is a binary tree structure commonly used for fast nearest-neighbor search. The leaf nodes represent individual VQ codes in the continuous embedding space of the tokenizer; each non-leaf node represents a high-dimensional ball. The root node represents a ball enclosing all VQ codes. Child nodes represent a smaller ball than the parent node. 

Concretely, given a $D$-depth ball tree with $2^{D-1}$ leaf nodes corresponding to tokens in $\mathcal{V}$, let  $N_d(c)$ be the set of all tokens that share a common ancestor with $c$ at depth $d$ where $1\le d \le D$.  Notably, $N_1(c)=\mathcal{V}$ is the whole vocabulary and $N_{D}(c)=\{c\}$ contains exactly one code.

We set $K=D-1$ as the total per-token corruption steps and associate them with $d=D-a$, with $a=0$ corresponding to $d=D$ and $a=K$ corresponding to $d=1$. Concretely, we generate $U_{j|k}$ with:
\begin{align}
    U_{k|k-1}(u|c)= \mathbf{1}\{u\in N_{D-k}(c) \}\frac{1}{|N_{D-k}(c)|},
\end{align}
which is a uniform probability vector over the set $N_{D-k}(c)$, i.e., we jump from code $c$ to a random neighboring token in the embedding space indicated by the ball tree, with the jump range dependent on the target token age. This process has the following marginal:
\begin{align}
    U_{j|k}(u|c)= \mathbf{1}\{u\in N_{D-j}(c) \}\frac{1}{|N_{D-j}(c)|},
\end{align}
for $j>k$ because the intermediate states are always leaf nodes of $N_{D-j}(c)$ by tree construction. An important observation is that for a fixed $u$, $U_{j|k}(u|c)$ is either 0 or $\frac{1}{2^j}$ for all possible $c\in\mathcal{V}$.

\begin{figure}
    \centering
    \includegraphics[width=1.0\linewidth]{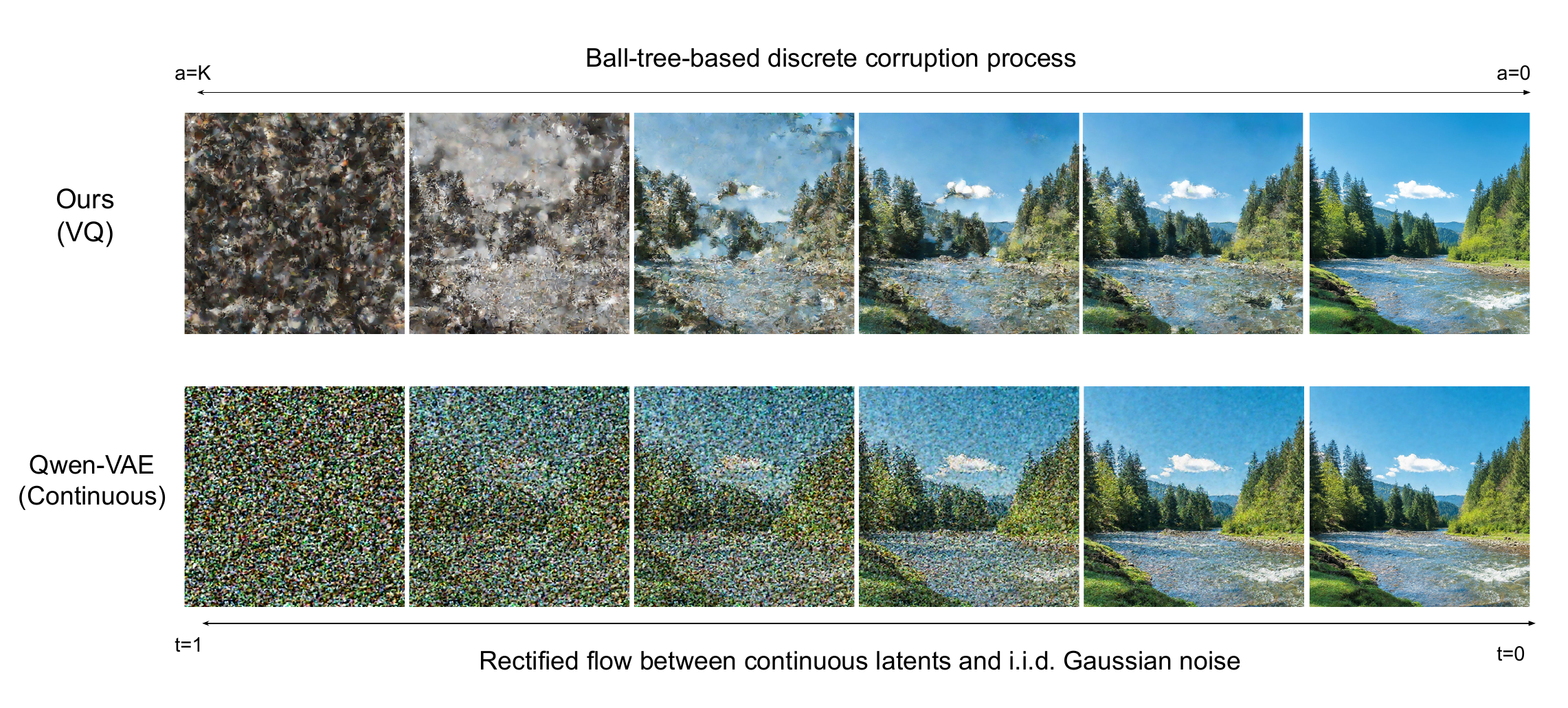}
    \caption{\textbf{Visual Reconstruction Across Different Token Ages.} To investigate whether the proposed age-based corruption process is semantically meaningful for visual generation tasks, we show reconstructed images at different token ages and compare them with the rectified flow process in the continuous latent space of Qwen-VAE. Results indicate that changes in the token age represent a smooth corruption process in the pixel space. In this visualization, all tokens are corrupted at the same rate, which corresponds to $\epsilon=1$. }
    \label{fig:appendix_corruption_vis}
\end{figure}

Let $\hat{c}$ be the initial token state at $a=0$. The conditional reverse process has the following form according to Bayes' rule:
\begin{align}
    U_{k|j}(c|u,\hat{c})= \frac{U_{k|0}(c|\hat{c}) U_{j|k}(u|c) }{ U_{j|0}(u|\hat{c})}.
\end{align}

Note that for this value to be properly defined, we must have $U_{j|0}(u|\hat{c})\ne 0$, implying $u\in N_{D-j}(\hat{c})$. When $U_{j|k}(u|c)\ne 0$, by the previously discussed observation the two terms cancel and the expression simplifies to $U_{k|0}(c|\hat{c})$.  When $U_{j|k}(u|c)= 0$, it can be implied that $u\notin N_{D-j}(c)$. Hence $N_{D-j}(c)\ne N_{D-j}(\hat{c})$, and consequently $N_{D-k}(c)\ne N_{D-k}(\hat{c})$ because $k$, which satisfies $k<j$,  corresponds to more refined tree nodes with greater depth. This implies that $U_{k|0}(c|\hat{c})=0$, so everywhere we have the reverse process satisfying:
\begin{align}
     U_{k|j}(c|u,\hat{c})= U_{k|0}(c|\hat{c}) =  \mathbf{1}\{c\in N_{D-k}(\hat{c}) \}\frac{1}{|N_{D-k}(\hat{c})|}.
     \label{eq:appendix_reverse}
\end{align}
An implementation detail worth mentioning is that a model $p_\theta$ may still predict $\hat{c}$ such that $u\notin N_{D-j}(\hat{c})$ as its output space is the entire vocabulary $\mathcal{V}$. In those cases, we still apply the RHS by allowing the model to correct its previous predictions. In practice, we find that \ours~rarely predicts a token that violates this setting, and the constrained setup shows no meaningful difference from the unconstrained setup.

Figure \ref{fig:appendix_corruption} illustrates an example of this process with $D=4$. On the left side, we visualize the set of leaf nodes $N_d(c)$ for $d=1,2,3,4$ corresponding to the token $c$ (encircled with a dashed line) in the embedding space, which are neighboring tokens to $c$ at different radii. On the right side, we show a binary tree representation of the same structure. The forward process starts with token age $a=0$ and ends with age $a=K$, which corresponds to traversing to parent nodes in the binary tree from the leaf node at $d=4$ to the root node at $d=1$. In the embedding space, it interpolates the one-hot distribution of a single embedding $c$ to progressively larger high-dimensional balls, until it reaches the full space.

\textbf{Empirical Implementation.} The tokenizer of \ours~contains 131k discrete codes, which corresponds to a ball tree with $D=18$. Hence, the vanilla setup is $K=17$.  To enable flexible inference, we in practice randomize $K$ for each training step among the candidates $\{1,2,..17\}$, where for $K<17$, we skip intermediate ages uniformly. For example, $K=4$ corresponds to age progression $0,4,8,12,17$ in the original $K=17$ notation. 

To verify that the ball-tree-based corruption process is semantically meaningful, we visualize the progression of corrupt tokens in Figure \ref{fig:appendix_corruption_vis} and compare it with the continuous rectified-flow setup. Specifically, we encode the images with \ours's tokenizer and visualize the codes sampled from the defined corruption process $U_{k|k-1}(u|c)$. We also encode the images with the continuous VAE of Qwen-Image \cite{wu2025qwen} and interpolate their latents with i.i.d. Gaussian noise following the rectified flow formulation. Visual results show that both processes gradually corrupt the semantics of the original image in a smooth manner. 

\begin{figure}
    \centering
    \includegraphics[width=0.8\linewidth]{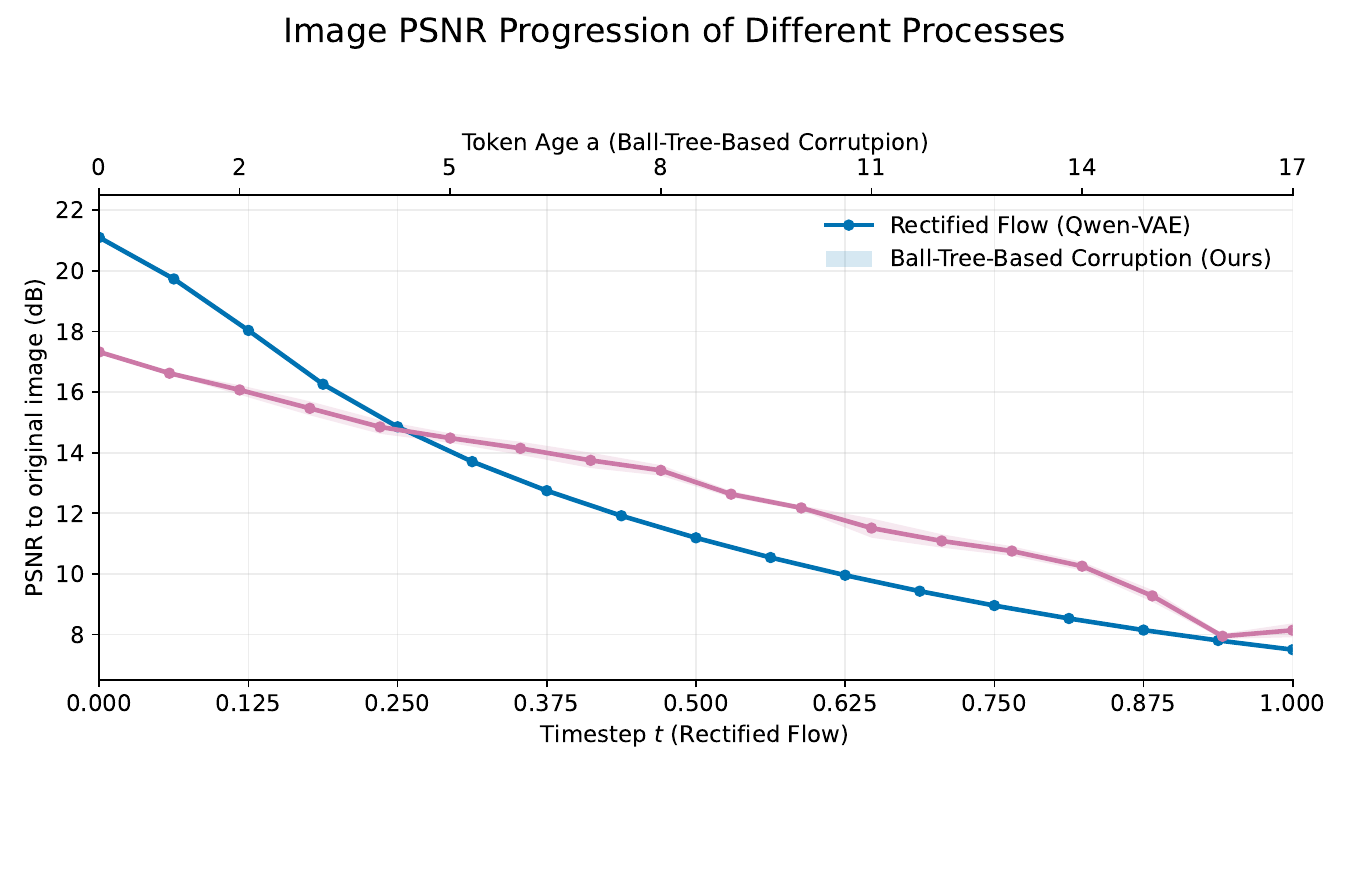}
    \caption{\textbf{Image PSNR Across Different Token Ages.} To investigate whether the proposed age-based corruption process is semantically meaningful for visual generation tasks, we plot the PSNR of reconstructed images at different token ages and compare them with the rectified flow process in the continuous latent space of Qwen-VAE. Results indicate that changes in the token age represent a smooth corruption process in the pixel space measured by PSNR. In the clean state, Qwen-VAE has better PSNR than \ours, suggesting its continuous VAE has better reconstruction capability.  }
    \label{fig:imag_psnr}
\end{figure}

We further confirm this observation quantitatively by measuring the PSNR against the ground truth image. The results are shown in Figure \ref{fig:imag_psnr}. The plot demonstrates that both processes offer a smooth transition from a clean image to random noise measured by PSNR. We observe that rectified flow has higher PSNR at zero corruption; this is attributed to the better reconstruction capability of Qwen-VAE as opposed to the corruption process.

As a final remark, we clarify that $U_{j|k}$ is a discrete-time per-token corruption process defined over discrete token ages $0,1...K$, it should not be confused with the continuous-time diffusion process $q(X^t|X^s)$ with continuous timesteps $t,s\in[0,1]$. The two processes are bridged by the token age field, which determines when discrete jumps $U_{k-1|k}$ occur over the continuous time interval $[0,1]$ for each token.

\subsubsection{Non-Markov Per-Token Jump Schedule} 
An important observation in the formulation described in Section \ref{sec:appendix_sbd} is that during training, only the marginal:
\label{sec:appendix_jump_schedule}
\begin{equation}
    q(X_{i}^t|X_{i}^0) = (1-t)\mathbf{X_{i}^0}+ t \textbf{M}, 
\end{equation}
is used to construct the partially masked sequence $X^t$ and we never explicitly invoke $q(X_{i}^t|X_{i}^s)$ from Equation \ref{eq:sbd_forward_transit} for $s\ne0$. Discrete non-Markov diffusion models (DNMD) \citep{chen2024fast} constructs a process that gives the same forward marginal with deterministic token transition time. Specifically, the forward process can be defined as:
\begin{equation}
    q(X_{i}^t|X_{i}^s,\tau_i) = \mathbf{1}\{t\le\tau_i\}\mathbf{X_{i}^s}+ \mathbf{1}\{t>\tau_i\}\ \textbf{M}, 
\end{equation}
where $\tau_i\sim U([0,1])$ is the clean-to-mask transition time that is pre-sampled once for each token and maintained throughout the trajectory.  This design relaxes the Markov constraint of the process.\footnote{For the single-jump construction with a distinct absorbing
mask token, the process remains Markov after marginalizing  $\tau$ because the current state reveals whether the jump has occurred. More general single-jump DNDM can be
non-Markov. For example, suppose the initial token is replaced by an independent
uniform draw from $\mathcal{V}$ at a presampled time $\tau$, and then held fixed.} However, it is compatible with standard MDM training because it does not change the forward marginal or the clean token target distribution $p(X^0_i|X^t)$. 

We make an additional remark that DNDM's setup in fact better aligns with the empirical sampling algorithms \cite{chang2022maskgit,besnier2025halton,li2025lavidao} which often predefine an unmasking schedule, whereas the exact Markov reverse process would unmask a stochastic number of tokens, or even zero tokens, at each step. Most recent samplers for masked image models like the Halton sampler \citep{besnier2025halton}, and the stratified sampler \citep{li2025lavidao} actually pre-compute the exact timestep to unmask each token, making DNDM a more accurate formulation in the image generation domain. 

To extend this setup to allow intermediate transition states between the clean state and the mask state, we augment the generating process by first sampling a predefined transition period $[\tau_{i},\tau_{i}']$ as opposed to a fixed jump time $\tau_i$. In particular, we sample 
\begin{align}
    \tau_{i}\sim U([0,1-\epsilon]), \tau_{i}' = \tau_{i}+\epsilon,
\end{align}
where $\epsilon\in(0,1]$ is the hyper-parameter controlling the length of the transition process from clean state with $a=0$ to the mask state with $a=K$. In particular, when $\epsilon\rightarrow0$, all jumps happen in an infinitesimal time interval and the overall process reduces to the canonical DNDM. 

With a pre-sampled interval, the token age $A_i$ can be expressed as a deterministic step function 
\begin{align}
    a_i(t,\tau_i,\epsilon)= \begin{cases}
        0  & \text{if} \quad 0 \le t \le \tau_i\\
         \text{Floor}(K\frac{t-\tau_i}{\epsilon})& \text{if} \quad \tau_i < t < \tau'_i\\
        K & \text{if} \quad \tau_i' \le t \le 1 \\
    \end{cases}
.
\end{align}
This induces a one-hot categorical distribution over the token age $A_i^t$:
\begin{align}
    b_i^t(a)=\mathbb{P}(A_i^t=a|\tau_i,\epsilon)=\mathbf{1} \{a=a_i(t,\tau_i,\epsilon) \}.
\end{align}
Hence, the forward process conditioned on a pre-sampled interval is:
\begin{align}
q\!\left(X_i^t=(u,a)\mid X_i^s=(c,a'),\tau_i\right)
=
b_i^t(a)
\begin{cases}
U_{a\mid a'}(u\mid c), & a>a',\\
\mathbf{1}\{u=c\}, & a=a',\\
0, & a<a'.
\end{cases}
.
\label{eq:self_refinement_forward}
\end{align}

The posterior reverse process is simply
\begin{align}
&p\!\left(
X_i^s=(c,a')
\mid X_i^t=(u,a),X_i^0=(\hat{c},0),\tau_i
\right) =
b_i^s(a')
\begin{cases}
U_{a'\mid a}(c\mid u,\hat{c}), & a'<a,\\
\mathbf{1}\{c=u\}, & a'=a,\\
0, & a'>a.
\end{cases}
,
\end{align}
where $U_{a|a'}(u|c)$ and $U_{a'|a}(c|u,\hat{c})$ are the forward and reverse marginals of the discrete-time per-token corruption process. At inference, $X_{i}^0$ is sampled from the model's clean token prediction.

\subsection{On-policy Distillation from AR VLMs}

\label{sec:appendix_opd_und_detail}

\newcommand{\pstudent}{p_\text{student}}
\newcommand{\pteacher}{p_\text{teacher}}
The standard OPD framework is defined over AR language models sharing the same vocabulary.  Specifically, given a teacher model and student model, the OPD loss \citep{gu2024minillm} is defined as:
\begin{align}
\mathcal{L_\text{AR-OPD}}=\mathbb{E}_{X^0\sim\pstudent}[\frac{1}{L}\sum_{i=1}^L\mathbb{D}_\text{KL}( \pstudent(X_i^0|X_{j<i}^0) || \pteacher(X_i^0|X_{j<i}^0)) ], 
\label{eq:opd_ar_loss}
\end{align}
where $X^0=[X^0_1...X^0_L]$ is a sequence sampled from the student model. Recall from Section \ref{sec:appendix_sbd} that the SBD model divides $X^0$ into consecutive blocks $X^0=[B^0_1,..B^0_K]$  and implements 
\begin{align}
    p_\theta(B_{b,i}^0|B^0_{j<b},B_b^t)
    \label{eq:block_prob}
\end{align}
where $B_b^t$ is a partially masked version of the clean block $B_b^0$ and $B_{b,i}^0$ is the $i$-th token in the $b$-th block.

When the block size is 1, each block only has two states: a clean token and a mask token, so it is generated in one step by evaluating
\begin{align}
    p_\theta(B_{b,1}^0|B^0_{j<b},B_b^1)=p_\theta(B_{b,1}^0|B^0_{j<b})=p_\theta(X_{b}^0|X^0_{j<b})
\end{align}
where $B_b^1$ can be marginalized since there is only one possible state with a probability of 1. The final expression exactly implements an AR model, and we refer to this setup as the AR view of the SBD model. Practically, this distribution can be evaluated for all tokens in a sampled sequence in a single pass because of the shift-by-one logit implementation illustrated in Figure \ref{fig:sbd_training}. Given this AR generation mode and AR view, we can directly apply the OPD loss used in AR models from Equation \ref{eq:opd_ar_loss} to the SBD model with no change.

However, if we only sample from and supervise the model in its AR view, the SBD model would lose its support for parallel generation with block sizes larger than 1. To provide generic supervision for Equation \ref{eq:block_prob} with a block size larger than 1, we define the AR-Diffusion Consistency loss as: 
\begin{align}
\mathcal{L_\text{Cons}}=\mathbb{E}_{X^0\sim\pstudent,t\sim U([0,1])}[\frac{1}{KNt}\sum_{b=1}^K\sum_{i=1}^N\mathbb{D}_\text{KL}( \pstudent(B_{b,i}^0|B_{j<b}^0,B_{b,r<i}^0) || \pstudent(B_{b,i}^0|B_{j<b}^0,B_b^t)) ]   ,
\label{eq:opd_cons_loss}
\end{align}
where $N>1$ is the block size and $B_b^t$ is a partially masked block sampled from the forward marginal $q(B_b^t|B_b^0)$. $X^0$ is sampled with AR mode with block size 1. The final loss for visual understanding and reasoning tasks is:
\begin{align}
\mathcal{L}_\text{OPD-UND} = \mathcal{L}_\text{AR-OPD} + \mathcal{L}_\text{Cons}.
\end{align}

\subsubsection{Connection with OPD-LM}

In our notation convention, OPD-LM \citep{su2026opdlm}'s objective can be represented as a single term:
\begin{align}
\mathcal{L_\text{OPD-LM}}=\mathbb{E}_{X^0,X^t\sim\pstudent}[\frac{1}{KN}\sum_{b=1}^K\sum_{i=1}^N\mathbb{D}_\text{KL}( \pteacher(B_{b,i}^0|B_{j<b}^0,B_{b,r<i}^0) || \pstudent(B_{b,i}^0|B_{j<b}^0,B_b^t)) ].
\end{align}

Compared with our consistency loss, the main differences are
\begin{itemize}
    \item OPD-LM samples rollouts from the diffusion view while \ours~samples from the AR view
    \item OPD-LM directly uses teacher scores to supervise the student's diffusion view, while \ours~uses the student's own AR score to supervise the diffusion view
    \item OPD-LM initializes the student with the teacher's weights, as an untrained diffusion model. In cross-paradigm OPD of \ours, the teacher and the student are two distinct models with different sizes and architectures. The student model already has strong diffusion generation capabilities before OPD. 
\end{itemize}  

At the system level, the broader difference is that \ours~uses OPD as a component in a multi-objective unified training run combining visual understanding, text-to-image, and image editing objectives, while OPD-LM focuses solely on single-objective AR2Diff post-training in the language domain. 

One of the problems with OPD-LM's setup is that it creates effectively ill-formed supervision signals due to potentially conflicting contexts. Specifically,  high-probability tokens in $\pteacher(B_{b,i}^0|B_{j<b}^0,B_{b,r<i}^0)$ may not be good candidates semantically for $\pstudent(B_{b,i}^0|B_{j<b}^0,B_b^t)$ with a different context.

As a concrete example,  consider the following student generation \texttt{"The number \(i\) is called the multiplicative identity."}. Given the prefix \texttt{"The number \(i\) is called the"}, the teacher may want to generate the term for \texttt{"imaginary unit"}, and assign high scores to the token \texttt{"imaginary"}, but it would not be a good candidate if the masked context is \texttt{"The number \(i\) is called the [m] identity"} because the term \texttt{"imaginary identity"} is ill-formed.

The consistency loss of \ours mitigates this issue by using the student's own AR predictions as supervision targets, avoiding the additional teacher--student distribution discrepancy while retaining the difference between AR and diffusion contexts. At zero sampling temperature, each token in a greedy rollout is an argmax of the student's AR conditional distribution, although this does not guarantee compatibility of the full supervision distribution with the observed future context. Empirically, we find that sampling AR rollouts at a small nonzero temperature yields better diffusion performance, suggesting that increased trajectory diversity can be beneficial despite relaxing the argmax property.

In Table \ref{tab:math-reasoning-results} of the main paper, we show that directly applying OPD-LM's loss with heterogeneous teacher and student models performs poorly because of the mismatch identified above. We provide more analysis of each individual component in Section \ref{sec:appendix_ablation_opd_und}.

\subsection{On-policy Distillation from continuous flow models}
 
\begin{figure}[h]
    \centering
    \includegraphics[width=0.8\linewidth]{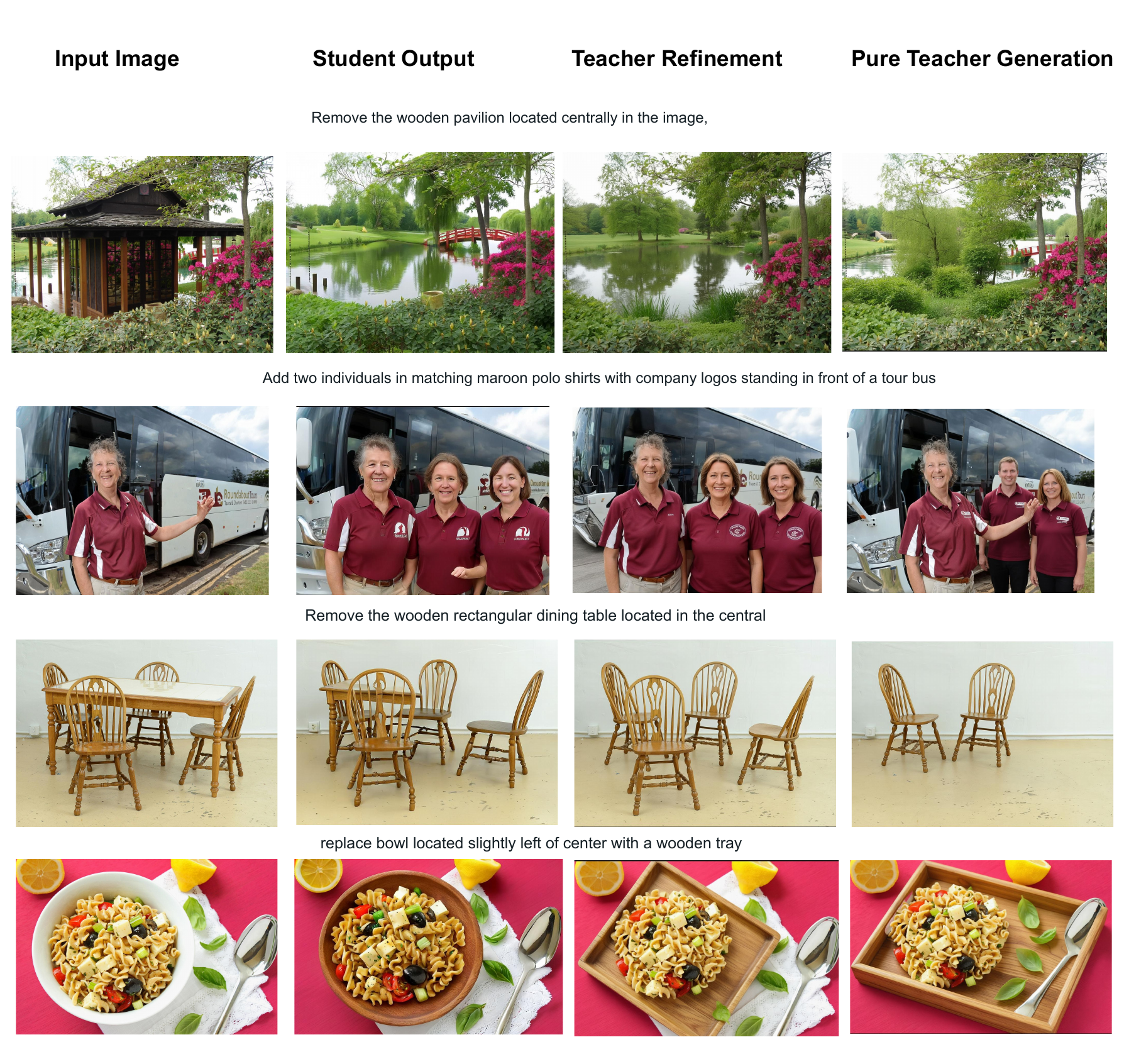}
    \caption{\textbf{Additional Qualitative Illustrations of the Teacher Refinement Process.} We show additional qualitative samples of teacher-refined student outputs. They generally align with the student roll-outs but have better visual fidelity and source image preservation. In some examples where the teacher over-edits, the refined student outputs are preferable to pure teacher generation. }
    \label{fig:refinevisappendix}
\end{figure}

\label{sec:opd_cont_appendix_method}

In distilling from Qwen-Image on image editing tasks, we start with an input image $V_\text{in}$ and text instructions $p$. We then generate discrete codes $X_{out}^0$ by sampling from the self-refining diffusion process $p_\theta(X_{\text{out}}^0|V_\text{in},p)$. We decode them with \ours's VQ decoder to obtain the image in pixel space $V_{out}=\text{Decode}_\text{VQ}(X_{\text{out}})$. We then re-encode $V_{out}$ with the teacher's continuous VAE $x^0=\text{Encode}_\text{VAE}(V_{\text{out}})$, and then create a partially noised version of the latent  $x^r=(1-r) x^0+r \delta$ where $\delta\sim N(0,I)$ is i.i.d. Gaussian noise and $r$ is a hyperparameter controlling refinement strength. The teacher model $\pteacher(\hat{x}^0|x^r,p,V_\text{in})$ is employed to sample the clean prediction $\hat{x}^0$. Crucially, this refinement process involves multiple ODE steps as opposed to a single-step teacher forward pass.

The predicted latent $\hat{x}^0$ is decoded into pixel space and re-encoded into discrete VQ codes to obtain $\hat{X}^0=\text{Encode}_\text{VQ}(\text{Decode}_\text{VAE}(\hat{x}^0))$. We then sample the partially corrupted sequence $\hat{X}^t$ using Equation \ref{eq:self_refinement_forward}. The final OPD objective is defined by:
\begin{align}
    \mathcal{L}_\text{OPD-Gen}(\hat{X}^0,\hat{X}^t,V_\text{in},p,t)=- \frac{1}{t}\log p_\theta(\hat{X}^0|\hat{X}^t,V_\text{in},p),
\end{align}
which is effectively the SFT loss on the refined target. The full process is described in Algorithm \ref{alg:opd_gen}.

\begin{algorithm}[h]
\caption{Teacher-Refined Distillation from On-Policy Student Rollouts}
\label{alg:opd_gen}
\begin{algorithmic}[1]
\Require Student $p_\theta$; frozen teacher $T$; training data $\mathcal D$;
         refinement strength $r\in(0,1)$; ODE steps $M>1$;
         corruption-time distribution $\rho$ on $(0,1]$
\While{not converged}
    \State Sample a minibatch $\mathcal B$ from $\mathcal D$
    \State $\mathcal L \gets 0$
    \For{$(V_{\mathrm{in}},p)\in\mathcal B$}
        \State $X_{\mathrm{out}}^0
            \sim p_\theta(\cdot\mid V_{\mathrm{in}},p)$
            \Comment{Current student rollout}
        \State $V_{\mathrm{out}}
            \gets \operatorname{Decode}_{\mathrm{VQ}}
            (X_{\mathrm{out}}^0)$
        \State $x^0
            \gets \operatorname{Encode}_{\mathrm{VAE}}
            (V_{\mathrm{out}})$
        \State $\delta\sim\mathcal N(0,I)$
        \State $x^r\gets(1-r)x^0+r\delta$
        \State $\hat x^0
            \gets \operatorname{ODESolve}_{T}
            (x^r;\,r\!\rightarrow\!0,\,M,\,V_{\mathrm{in}},p)$
            \Comment{Multi-step teacher refinement}
        \State $\hat X^0
            \gets \operatorname{stopgrad}\!\left(
            \operatorname{Encode}_{\mathrm{VQ}}\!\left(
            \operatorname{Decode}_{\mathrm{VAE}}(\hat x^0)
            \right)\right)$
        \State Sample $t\sim\rho$
        \State Sample $\hat X^t$ by corrupting $\hat X^0$
            using Eq.~\ref{eq:self_refinement_forward}
        \State $\mathcal L\gets\mathcal L
            -\dfrac{1}{t}
            \log p_\theta
            (\hat X^0\mid\hat X^t,V_{\mathrm{in}},p)$
    \EndFor
    \State Update $\theta$
\EndWhile
\end{algorithmic}
\end{algorithm}

\textbf{Connection to Trajectory-refined Distillation in Language Modeling.} We note that while our objective is an SFT-style loss, the learning process is on-policy because the supervision target is derived from the student output. This paradigm closely connects with several concurrent works in the OPD literature in the language domain that explored student-teacher interleaved generation.  Notably, Trajectory-refined Distillation \citep{jiang2026trajectory} proposes to use a teacher model to correct student trajectories in language tasks. \ours~expands this concept across model families and to the visual generation domain to address a lack of strong MDM teachers in visual generation tasks. 

\textbf{Qualitative Assessment.} In addition to Figure \ref{fig:distillaion}, we provide additional samples in Figure \ref{fig:refinevisappendix} to illustrate the effectiveness of our proposed approach. We make two observations. First, compared with pure teacher generation, the teacher-refined student outputs are visually aligned to the student output but with more refined details and better source-image preservation, making them a more ideal training target. Second, we find that the teacher-refined student output can actually be better than the student model's outputs when the pure teacher generation makes a mistake. In the example \texttt{"Remove the wooden rectangular dining table"}, teacher-refined student generation achieves the best results and outperforms both student and teacher generation. This occurs because when the teacher makes a mistake, the pure teacher generation tends to drift from the student example considerably (in this case, over-editing and removing two extra chairs in addition to the table). The refinement process, which constrains the teacher output to a neighborhood of the student roll-out, serves as useful regularization in these cases.

\section{Additional Details of the Main Training Run}
\label{sec:appendix_training_setup}
\subsection{Data Mixture}

Our training data consists of the following components: 

\begin{itemize}
    \item \textit{A: Text-to-Image Pairs.} Image sources include LAION-2B \citep{schuhmann2022laion}, COYO-700M \citep{kakaobrain2022coyo-700m}, SA-1B \citep{kirillov2023segment}, JourneyDB \citep{sun2023journeydb}, BLIP3o-60k \citep{chen2025blip3}, ShareGPT4o-Image \citep{chen2025sharegpt}. The images are filtered to remove NSFW prompts, watermarks, low CLIP scores \citep{radford2021learning}, low aesthetic scores \citep{laion-aesthetics}, and low-resolution samples. This results in 127M images in the final mix. Images are recaptioned with Qwen3-VL \citep{bai2025qwen3} when only alt-text is available.
    \item \textit{B: Visual Instruction Tuning Data.} We include LLaVA-OneVision \citep{li2024llava}, Open-LLaVA-Next \citep{chen2024open}, MAmmoth-VL \citep{guo2024mammoth}, and VisualWebInstruct \citep{visualwebinstruct}.  We additionally include grounding-specific data including GranD \citep{hanoona2023GLaMM} and RefCOCO \citep{kazemzadeh2014referitgame}. We additionally include the entire training set of Nemotron-3-Nano-Omni \cite{deshmukh2026nemotron}.
    \item \textit{C: Image Editing Data.} We include ShareGPT4o-Image \citep{chen2025sharegpt}, GPT-Edit-1.5M \citep{wang2025gpt}, the image editing subset of UniWorld-V1 \citep{lin2025uniworld}, and ScaleEdit-12M \citep{chen2026scaleedit}.
\end{itemize}

The OPD dataset is sampled from the prompts of the full dataset with labels discarded. Specifically, for visual reasoning OPD, we select 76.8k samples from the datasets in part (B) prioritizing visual reasoning tasks. For image editing tasks, we randomly select 100k samples from datasets in part (C).

\subsection{Optimization Setup}

We use NLD \citep{fu2026nemotron} to initialize the understanding branch and semantic encoder. For the VQ encoder, we adopt a pretrained IBQ \citep{cui2025emu3}. For the semantic encoder, we adopt RADIO-H \citep{heinrich2025radiov2}. The image generation branch is initialized from the copied weights of the understanding branch. The overall two-tower structure follows Elastic-MoT design proposed by LaViDa-O \citep{li2025lavidao}.

Training consists of three stages. In the first stage, we fine-tune the language model on image-text pairs from dataset (A).  In the second stage, we add instruction data and image editing data from (B) and (C). In the final stage, we use the OPD dataset described above. Since we do not perform OPD on T2I tasks, we mix image-text pairs from (A) to preserve text-to-image ability. 

In the main paper, the checkpoint after stage two is referred to as \ours-Base. The final checkpoint is referred to as \ours. The full details of the optimization are listed in Table \ref{tab:training-stages}.

\begin{table}[t]
\centering
\caption{\textbf{Training configurations across three stages.} We use letters A-C to represent different dataset. }
\label{tab:training-stages}
\begin{tabular}{lccc}
\toprule

 & \textbf{Stage 1} & \textbf{Stage 2} & \textbf{Stage 3} \\
\midrule
Learning Rate & $1 \times 10^{-5}$ & $1 \times 10^{-5}$ & $2 \times 10^{-6}$ \\
Steps & 400k & 100k $\times$ 2 & 10k \\
$\beta_1$ & 0.99 &0.99  & 0.99 \\
$\beta_2$ & 0.999 &0.999  & 0.999 \\
optimizer & AdamW & AdamW & AdamW \\
\midrule
Dataset Used & A & A,B,C & B and subset of A, C  \\
Objective & SFT & SFT & SFT+OPD  \\
\bottomrule
\end{tabular}%

\end{table}

\subsection{Infrastructure}

The training is conducted on 64 H100 GPUs across 8 nodes over 26 days. In particular, the final OPD stage takes 5 days. 

\section{Additional Experiment Results}
\label{sec:appendix_ablation}
\subsection{Studies on the self-refining diffusion process}

\subsubsection{Comparison with MDM-Prime}

\label{sec:appendix_mdm_prime}
\begin{figure}
    \centering
    \includegraphics[width=1.0\linewidth]{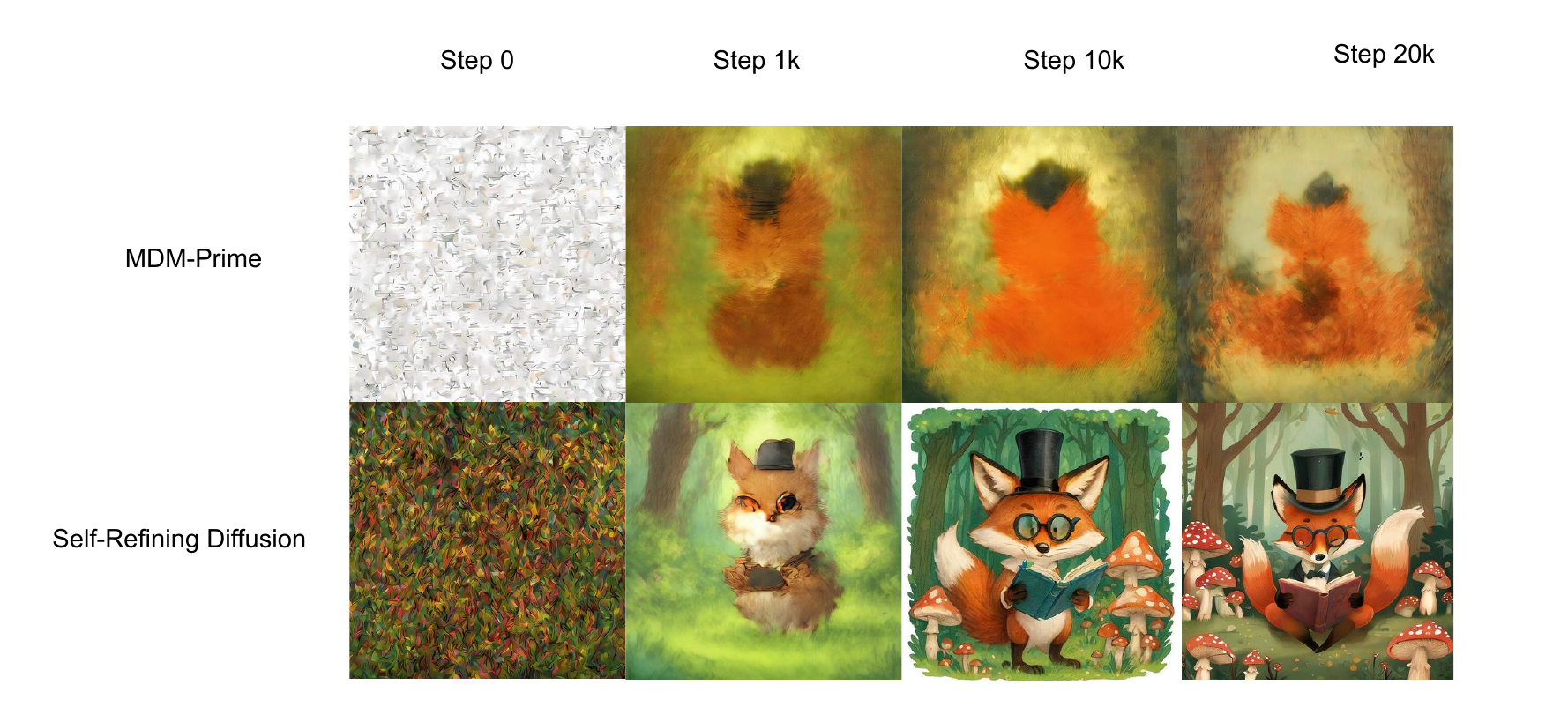}
    \caption{\textbf{Comparison with MDM-Prime.} Progression of Generated Images at Different Training Steps. The prompt is \textit{"Highly stylized, whimsical illustration of a fox wearing a tiny monocle and top hat, reading a miniature book in a cozy, mushroom-filled forest glade. Children's book art, gentle colors, charming, adorable."}}
    \label{fig:mdm_prime}
\end{figure}

We additionally tried to compare against a related method, MDM-Prime \cite{chao2026beyond} which also uses partial masking, but it shows no sign of convergence with a score of 0 on GenEval even after 20k steps of training. The progression of generated images throughout training is shown in Figure \ref{fig:mdm_prime}. 

Unlike our self-refining process which is shown to correspond to a smooth corruption in pixel space in Figure \ref{fig:appendix_corruption_vis} and Figure \ref{fig:imag_psnr}, MDM-Prime's setup simply factorizes a token index $v\in \mathcal {V}$ into sub-tokens  $(v_1,v_2,..v_l)\in \mathcal {V}_1 \times \mathcal {V}_2...\mathcal {V}_l$, with $|\mathcal{V}_i|=|\sqrt[l]{|\mathcal{V}|}|$.

For example, with a vocabulary of 16 tokens and $l=4$, token 3 is represented as \texttt{0011}. The intermediate mask states simply mask the bits, giving states like \texttt{0[m][m]1} or \texttt{001[m]}. To represent such states, it learns a per-sub-token embedding and concatenates them to form the token embedding. This design materially reduces the total parameters of the embedding table and reduces its expressive power. In MDM-Prime's visual generation setup, it sets $|\mathcal{V_i}|=16$ which means some dimensions of the token embeddings corresponding to codes $1,17,33,...$ are exactly identical, which can be limiting in large-scale T2I settings because these tokens are not guaranteed to have any similarity at all. We note that MDM-Prime's visual generation experiments are conducted on tiny $32\times32$ images in CIFAR-10 \citep{krizhevsky2009learning} and ImageNet-32 \citep{deng2009imagenet}.

We additionally note that while the training failed to converge at $1024\times1024$ resolution, MDM-Prime's generated images still show some high-level object-like structure, as shown in Figure \ref{fig:mdm_prime}, which could explain why it can succeed on $32\times32$ low-resolution images.

\subsection{Effect of refinement steps}

While in the formulation we use $\epsilon\in[0,1]$ to define the progression speed of intermediate states in self-refining diffusion, what matters in practical inference settings is the total number of inference steps $\text{NFE}$ and the number of refinements each sampled token goes through, denoted by $N$. For example, $N=1$ corresponds to vanilla mask diffusion with $\epsilon\rightarrow0$, while $N=\text{NFE}$ means all tokens follow the same refinement schedule across the entire sampling process, or $\epsilon=1$.  $N>\text{NFE}$ is not possible by definition.

We show the interaction effect of $\text{NFE}$ and $N$ in Table \ref{tab:hpsv3_nfe_k} at inference time. We note that at any total NFE, setting $\epsilon=1$ works poorly. At small NFE, $N=1$ which corresponds to a single state jump works best. By contrast, at larger NFE, a higher number of refinement steps starts to show benefits, with $N=8$ working best in our default inference setup with 64 steps. Increasing N from 8 to 16 provides no further improvement in the tested settings and can substantially degrade performance at low NFE, where it implies an $\epsilon$ value closer to one. This suggests 8 refinement steps are sufficient. The failure patterns at low NFE and around $\text{NFE}=N$ suggest we should enforce some form of sequential order between the unmasking/refining processes of different tokens, as opposed to unmasking/refining a large number of tokens simultaneously, which can be challenging.

\begin{table}[t]
\centering
\caption{\textbf{Effect of Refinement Steps Per Token.} We report HPSv3 scores across actual NFE and number of refinement stages $N$.}
\label{tab:hpsv3_nfe_k}
\begin{tabular}{rcccc}
\toprule
\textbf{Actual NFE} & $\boldsymbol{N=1}$ & $\boldsymbol{N=4}$ &
$\boldsymbol{N=8}$ & $\boldsymbol{N=16}$ \\
\midrule
4  &\textbf{ 7.3700}  & -2.4895          & \textemdash & \textemdash \\
8  & \textbf{9.6214}  & 9.0038           & 0.0834      & \textemdash \\
16 & 10.5449 & \textbf{10.8205} & 10.3044     & 3.0348      \\
32 & 10.8921 & \textbf{11.0137} & 10.9894     & 10.7620     \\
48 & 10.9378 & 11.0369          & \textbf{11.1167} & 11.0861 \\
64 & 11.0051 & 11.0287      & \textbf{11.1050} & 11.1010 \\
\bottomrule
\end{tabular}
\end{table}

\subsection{Studies on distilling from AR VLMs}
\label{sec:appendix_ablation_opd_und}

We conducted additional ablations on the proposed OPD framework. These results are initialized with \ours-Base model and post-trained only on multi-modal reasoning tasks. The results are not directly comparable to the main result which reports unified multi-task joint training.

\label{sec:appendix_temperature_ablation}
\subsubsection{Training Objectives}
\begin{figure}
    \centering
    \includegraphics[width=1.0\linewidth]{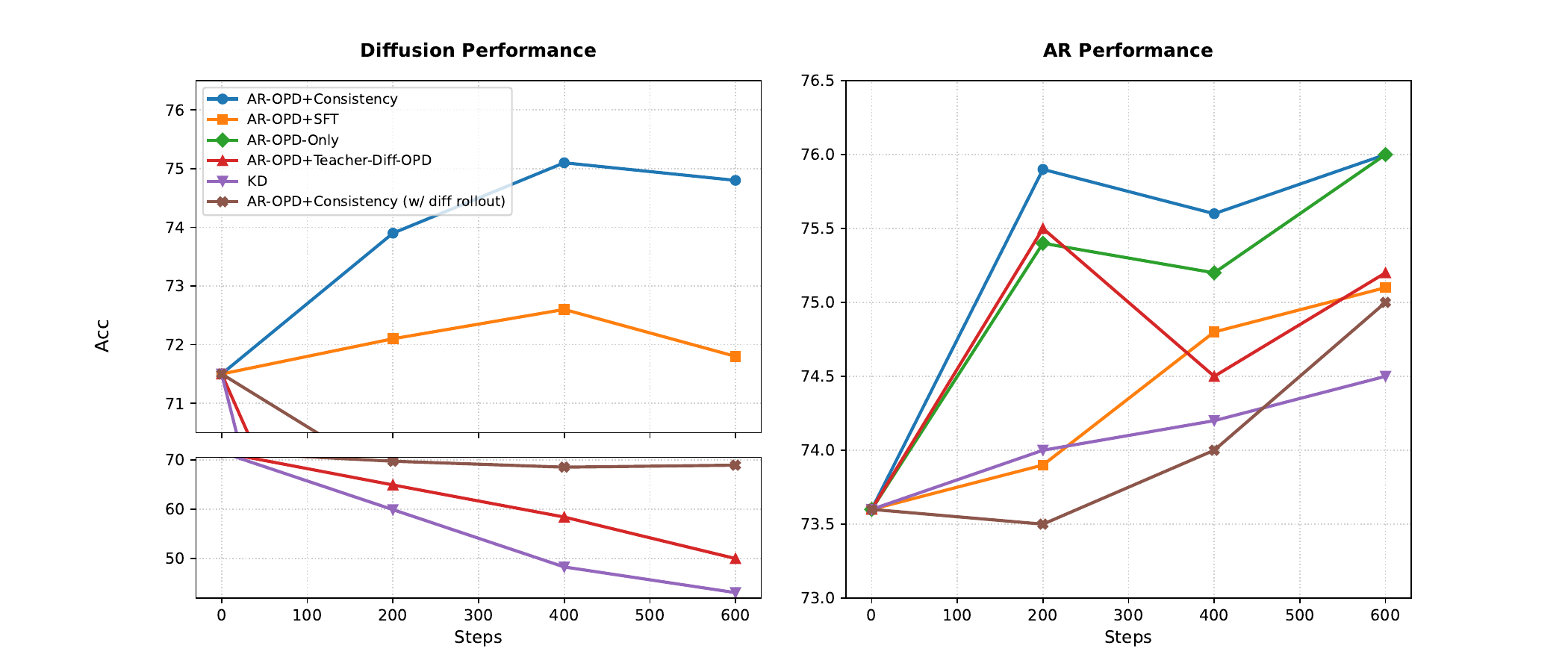}
    \caption{\textbf{Ablation Studies on distilling from AR VLMs}. We report MathVista Acc throughout training steps.}
    \label{fig:opd_ablation}
\end{figure}

We conducted detailed ablations on each component of the proposed OPD framework. Specifically, we compare the following variants of the method: \begin{itemize}
    \item (1) AR-OPD+Consistency:  The final chosen setup with AR OPD loss and consistency loss. \item (2) AR-OPD+SFT: replace consistency loss with a one-hot SFT objective on the diffusion view, using AR-generated rollouts as targets.
    \item (3) AR-OPD only: which treats the SBD model as a pure AR model.
    \item (4) AR-OPD+Teacher-Diff-OPD: directly copy the teacher score as the supervision signal for the diffusion view.
    \item (5) KD: Replace student's AR rollout with pre-generated teacher diffusion rollout.
        \item (6) AR-OPD+Consistency (w/ diff rollout): Replace student's AR rollout in (1) with student's diffusion rollout.
\end{itemize} 

We evaluate these settings in both AR generation mode and diffusion generation mode. The results are reported in Figure \ref{fig:opd_ablation}. We draw the following two conclusions:

\textbf{AR rollouts support effective diffusion supervision.} In diffusion generation mode, only (1) and (2) improve upon the baseline because all other setups create mismatched supervision signals caused by mismatched context. In particular, even the student's own diffusion generation fails because the consistency loss uses the student's AR score as the target, which may predict high-probability tokens that are inconsistent given future context generated from the student;s diffusion distribution. This issue is mitigated when the sampled sequences and the supervision target both come from the student's AR distribution.

\textbf{Consistency Loss does not hurt AR generation.} While AR generation is not the focus of this work, we find that the AR performance with AR-OPD+diffusion loss is comparable to AR-OPD-Only, suggesting adding supervision to diffusion view does not negatively affect AR view's performance.

Beyond the above experiments, we also explored a run with consistency loss only. This run quickly collapses to an accuracy of 24.1 at 400 steps, which is outside the plotted range. For presentation clarity, it is not shown in the figure. However, this behavior is not surprising since it ties the outputs of the student's own generation modes without external supervision, which quickly leads to degenerate sequences. 

\subsubsection{Sampling Temperature}

We vary the temperature used to generate student AR rollouts during OPD and evaluate the resulting models in both diffusion and AR decoding modes. As shown in Table~\ref{tab:opd_temperature}, greedy rollouts achieve the highest AR accuracy (76.9\%), whereas a temperature of 0.4 achieves the highest diffusion accuracy (74.8\%). Thus, requiring each rollout token to be an argmax of the student's AR conditional distribution is not necessary for effective transfer to the diffusion view. The improvement at moderate temperatures suggests that greater rollout diversity can benefit diffusion training. Across all tested temperatures, AR-OPD+Consistency achieves higher diffusion accuracy than the alternative designs evaluated in our ablations, such as AR-OPD+SFT. Together, these results support using the student's own AR predictions to supervise its diffusion view, while indicating that strict per-token argmax alignment is unnecessary.

We clarify that all evaluations are performed with a temperature of 0 for consistency and reproducibility. Empirically, even for models with non-zero rollout temperature, using a zero temperature still gives better accuracy.
\subsubsection{Choice of Teachers}

In this section, we explore using an alternative teacher model Kimi K-2.6 \citep{moonshot_kimi_k26}  and compare with using Nemotron3-Nano-Omni \cite{deshmukh2026nemotron} teacher. Experiment results are shown in Table \ref{tab:union_opd_step1000}. Overall, Nemotron Teacher outperforms Kimi Teacher in our OPD setting, despite Kimi Teacher being the stronger model. This can be attributed to a larger teacher-student gap. In particular, unlike Nemotron teacher, Kimi teacher uses a different tokenizer and a different format of thinking traces. We note that these experiments are part of our early exploration and are based on an earlier checkpoint than \ours-Base. Hence results in  Table \ref{tab:union_opd_step1000} are not directly comparable to other ablation studies (E.g. Figure \ref{fig:opd_ablation} ).

\begin{table*}[t]
\centering
\small
\caption{\textbf{Ablation study of teacher choices.} Results are sampled in AR view with a block size of one.}
\setlength{\tabcolsep}{3pt}
\begin{tabular}{lccccccc}
\toprule
& AI2D & ChartQA & DocVQA & TextVQA & OCRBench & MMMU & MathVista \\
& Test & Test & Val & Val & & Dev/Val & Mini \\
\midrule
\multicolumn{8}{l}{\textit{Teacher Performance}} \\
Nemotron Teacher
& \textbf{88.44} & 89.80 & 95.73 & \textbf{85.06}
& 88.20 & 70.67 & 82.70 \\
Kimi Teacher
& 85.14 & \textbf{94.72} & \textbf{97.46} & 83.93
& \textbf{92.70} & \textbf{77.44} & \textbf{88.90} \\
\midrule
\multicolumn{8}{l}{\textit{\ours~ w/ OPD}} \\
OPD w/ Nemotron
& \textbf{85.98} & \textbf{90.28} & \textbf{95.01}
& \textbf{77.82} & \textbf{85.30} & \textbf{61.11}
& \textbf{78.10} \\
OPD w/ Kimi
& 83.91 & 78.00 & 92.49 & 76.76 & 80.80 & 52.33 & 71.60 \\
\bottomrule
\end{tabular}
\label{tab:union_opd_step1000}
\end{table*}

\begin{table}[t]
    \centering
    \caption{\textbf{Effect of OPD rollout temperature.}
    We vary the sampling temperature of student AR rollouts during
    OPD and report MathVista accuracy.}
    \label{tab:opd_temperature}
    \begin{tabular}{ccc}
        \toprule
        \textbf{Temperature} & \textbf{Diffusion Performance} & \textbf{AR  Performance}\\
        \midrule
        0.0 & 73.9 & \textbf{76.9} \\
        0.4 & \textbf{74.8} & 76.0 \\
        0.6 & 74.6 & 75.4 \\
        1.0 & 73.8 & 75.1 \\
        \bottomrule
    \end{tabular}
\end{table}

\subsection{Studies on distilling from continuous flow models}

We study the effect of noise strength $r$ in the teacher refinement process of image editing OPD with the Qwen-Image teacher. We report benchmark performance in Table \ref{tab:ablation_edit}. Among all tested values, $r=0.9$ works best. We note that the added Gaussian noise has a highly non-linear effect on the final output. Even with a high noise rate at $r=0.9$, the generated images are still aligned to the student roll-out and can be very different from a purely teacher-generated sample, as is shown in Figure \ref{fig:refinevisappendix}.  

\begin{table}[h!]
    \centering
    \caption{\textbf{Ablation on Noise Ratio}. Image Editing Results across $r$ values}
    \label{tab:ablation_edit}
    \vspace{0.5em}
    \begin{tabular}{l c c c  c}
        \toprule
        \textbf{Method} & $\mathbf{r=0.5}$ & $\mathbf{r=0.8}$ & $\mathbf{r=0.9}$  & $\mathbf{r=1.0}$ \\
        \midrule
        ImgEdit & 4.06 & 4.09 & 4.17  & 3.90 \\
        \bottomrule
    \end{tabular}
\end{table}

\subsection{Speed-Quality Tradeoff}

We examine the speed-quality tradeoff of \ours by measuring the tokens per forward pass, throughput (tokens per second), and answer accuracy on the MathVista benchmark. Speed-quality trade-off is achieved by tuning the confidence threshold used in the unmasking sampler. Since our model also has an AR view, we report AR's performance. Results are reported in Table \ref{tab:speed_math}. In particular, the vanilla setup offers 552.58 tokens/s compared to an AR baseline of 202.74 tokens/s. If we relax the threshold, \ours~can achieve up to 997.95 tokens/s with minor performance degradation.

\begin{table}[h]
    \centering
    \caption{\textbf{Speed-Quality Tradeoff on MathVista Benchmark.} Results are measured on H100 GPUs.}
\begin{tabular}{lrrr}
\hline
\textbf{Mode / Threshold} & \textbf{Tokens/Forward} & \textbf{Tokens/s/GPU} & \textbf{MathVista (\%)} \\
\hline
AR             & 1.00 & 202.74 & \textbf{76.1} \\
Diffusion 0.95 & 3.94 & 463.84 & 75.2 \\
Diffusion 0.90 & 4.64 & 552.58 & 75.0 \\
Diffusion 0.60 & \textbf{8.31} & \textbf{997.95} & 73.4 \\
Diffusion 0.50 & 7.23 & 734.03 & 70.8 \\
\hline
\end{tabular}
\label{tab:speed_math}
\end{table}

\section{Additional Related Works}
\subsection{Unified Architecture.}

\label{sec:related_unified}

Unified MLLMs are capable of both understanding and generation tasks. Integrating different capabilities into a single model presents unique architectural challenges. Generally, unified MLLMs have two common architecture choices. The first type, such as Janus~\citep{chen2025janus}, MMaDa~\citep{yang2025mmada}, Show-O\citep{xie2024show}, uses a single dense transformer to output both image and text tokens. The second type, such as BAGEL~\citep{deng2025emerging}, MetaQueries~\citep{pan2025transfer} and LaViDa-O~\citep{li2025lavidao} employs separate parameter sets for handling image and text modalities. In particular, mixture-of-transformers (MoT) architecture \citep{liang2024mixture} incorporates two separate sets of attention and FFN parameters to process different modalities, while still allowing cross-modal interaction via a joint attention mechanism. Under this paradigm,  several works such as X-Fusion and LM-Fusion \citep{mo2025x,shi2024lmfusion} further investigated architecture designs and training recipes of MoT. In particular, LaViDa-O \citep{li2025lavidao} proposes Elastic-MoT design, which decouples the attention layer in the final half of the model and only enables joint attention in early layers. Given an 8B base model, Elastic-MoT extends it to a 16B MoT. However, when performing understanding tasks, only 8B parameters are loaded. When performing generation tasks, only 8B generation parameters + 4B understanding parameters are loaded. When performing interleaved tasks, all 16B parameters are loaded. This elasticity makes the model particularly efficient and introduces no overhead compared to an 8B VLM when performing understanding tasks.

\section{List of Benchmarks}.

In this section, we provide a full list of benchmark used in this work and their source references.

\begin{itemize}
    \item \textbf{Generic Multi-modal Understanding}: MMMU \citep{yue2023mmmu},RealWorldQA \citep{xai2024realworldqa}, ChartQA \citep{masry-etal-2022-chartqa}, DocVQA \citep{mathew2021docvqa}, InfoVQA \cite{mathew2022infographicvqa}, TextVQA \citep{singh2019towards}, AI2D \citep{kembhavi2016diagram}, RefCOCO \citep{kazemzadeh2014referitgame}, ScienceQA \citep{lu2022learn}, ScreenSpot-Pro \citep{li2025screenspot}, OCRBench-V2 \citep{fu2026ocrbench}.
    \item \textbf{Multi-modal Reasoning}:  \citep{lu2023mathvista}, \citep{zhang2024mathverse}, \citep{wang2024measuring}, DynaMath \citep{zou2025dynamath}, Geometry3k \citep{lu2021inter}, LogicVista \citep{xiao2024logicvista}, We-Math \citep{qiao2025we}.

    \item \textbf{Multi-modal Generation}: GenEval \citep{ghosh2023geneval}, DPG \citep{hu2024equipdpg}, HPSv3 \cite{ma2025hpsv3}, ImgEdit \cite{ye2025imgedit}.
\end{itemize}

\section{Limitations and Future Work}

In this work, we propose \ours, introducing a self-refining diffusion process and cross-paradigm on-policy distillation. Despite considerable success, some limitations remain. First, while the cross-paradigm OPD framework allows us to distill an AR teacher to a diffusion student, it still requires the vocabulary spaces of the two models to match. In our setup, both models use the Ministral 3 \citep{liu2026ministral} tokenizer. Many existing works have explored cross-tokenizer distillation between AR models \cite{niu2026tokenizer}. We leave extending these methods to support cross-paradigm distillation to future work. Second, while \ours~achieves state-of-the-art performance among unified diffusion MLLMs, there may be considerable room for further optimization on the model architecture side. We leave that for future work as well since it is not the focus of this work. Third, we define the self-refining diffusion process with the non-Markov formulation with predefined per-token corruption intervals and uniform progression of intermediate states. This design is mostly motivated by minimizing complexity. Future work may explore defining a Markov process with a stochastic transition rate that allows tokens to stay at different intermediate states for varying lengths of time.  Finally, from a system perspective, \ours~lacks fine-grained text rendering capabilities because such data is not widely present in the training mix. We will address this issue in future by curating additional text-rendering-focused data.

\end{document}